\documentclass[journal,twocolumn]{IEEEtran}

\usepackage[linesnumbered,ruled,vlined]{algorithm2e}
\usepackage{hyperref}
\usepackage{url}
\usepackage{graphics} 
\usepackage{times} 
\usepackage{booktabs}       
\usepackage{amsfonts}       
\usepackage{nicefrac}       
\usepackage{microtype}      
\usepackage{mathtools}
\usepackage{graphicx}
\usepackage{cases}
\usepackage{array}
\usepackage{color}
\usepackage{float}
\usepackage{amssymb}
\usepackage{wrapfig}
\usepackage{subfigure}
\usepackage{amsmath}
\usepackage{soul}
\usepackage{amsthm}
\usepackage{graphicx} 
\usepackage{subfigure}
\usepackage{epstopdf}
\usepackage{algpseudocode}
\usepackage{xcolor}

\theoremstyle{definition}

\theoremstyle{remark}

\begin{document}
\title{\LARGE \bf
RUSSO: Robust Underwater SLAM with Sonar Optimization against Visual Degradation}

\author{Shu Pan$^{1}$, Ziyang Hong$^{1}$, Zhangrui Hu$^{1}$, Xiandong Xu$^{2}$, Wenjie Lu $^{1}$ and~Liang~Hu$^{1*}$,~\IEEEmembership{Senior Member,~IEEE}
\thanks{* Corresponding Author.}
\thanks{$^{1}$ S. Pan, Z. Hong, Z. Hu, W. Lu and L.~Hu are School of Mechanical Engineering and Automation, Harbin Institute of Technology, Shenzhen,
China. For correspondence: ~l.hu@hit.edu.cn.}

\thanks{$^{2}$ X. Xu is with School of Electrical and Information Engineering, Tianjin University, 92 Weijin Road, Tianjin 300072, China }
}

\markboth{Manuscript}%
{Shell \MakeLowercase{\textit{et al.}}: Bare Demo of IEEEtran.cls for IEEE Journals}

\maketitle

\begin{abstract}
Visual degradation in underwater environments poses unique and significant challenges, which distinguishes underwater SLAM from popular vision-based SLAM on the ground. In this paper, we propose RUSSO, a robust underwater SLAM system which fuses stereo camera, inertial measurement unit (IMU), and imaging sonar to achieve robust and accurate localization in challenging underwater environments for 6 degrees of freedom (DoF) estimation. During visual degradation, the system is reduced to a sonar-inertial system estimating 3-DoF poses. The sonar pose estimation serves as a strong prior for IMU propagation, thereby enhancing the reliability of pose estimation with IMU propagation. Additionally, we propose a SLAM initialization method that leverages the imaging sonar to counteract the lack of visual features during the initialization stage of SLAM. We extensively validate RUSSO through experiments in simulator, pool, and sea scenarios. The results demonstrate that RUSSO achieves better robustness and localization accuracy compared to the state-of-the-art visual-inertial SLAM systems, especially in visually challenging scenarios. To the best of our knowledge, this is the first time fusing stereo camera, IMU, and imaging sonar to realize robust underwater SLAM against visual degradation. We will release our open-source codes, dataset and experiment videos for the community\footnote[3]{Project website: https://github.com/CLASS-Lab/RUSSO}. 
\end{abstract}


\section{Introduction}\label{sec:intro}
The underwater simultaneous localization and mapping (SLAM) technique is essential for various underwater tasks such as environmental monitoring, offshore infrastructure inspection, ocean exploration, and search and rescue operations \cite{xanthidis2020navigation, yang2022monocular, 9204397}, as it provides underwater vehicles accurate pose estimates and detailed mapping in the surroundings. The underwater environment brings unique challenges to SLAM systems that rarely arise on the ground or the air, such as the unavailability of GPS, rapid illumination variations due to light attenuation, and the lack of structures and features in open water. To address such challenges, multimodal sensor fusion strategies are widely used in existing underwater SLAM methods. 

Cameras and sonars are typical sensors used in various underwater SLAM systems, supplemented by additional sensors such as DVLs, IMUs, and Bathymeters. The rich visual information provided by the camera makes it a key component of the visual SLAM system. Unfortunately, the vision-only SLAM system is not robust to adverse underwater environments. For instance, image quality can severely degrade in turbid water, and visual features are diminished in open water lacking structures or textures. A robust underwater SLAM system against visual degradation was proposed in  \cite{xu2021underwater, huang2023tightly}, where  DVL measurements were fused with camera images. However, the velocity and direction measurements of DVLs are directly influenced by water flow, which limits the localization accuracy in challenging water environments. 

\begin{figure}[t!]
  \centering
  \includegraphics[width=0.48\textwidth]{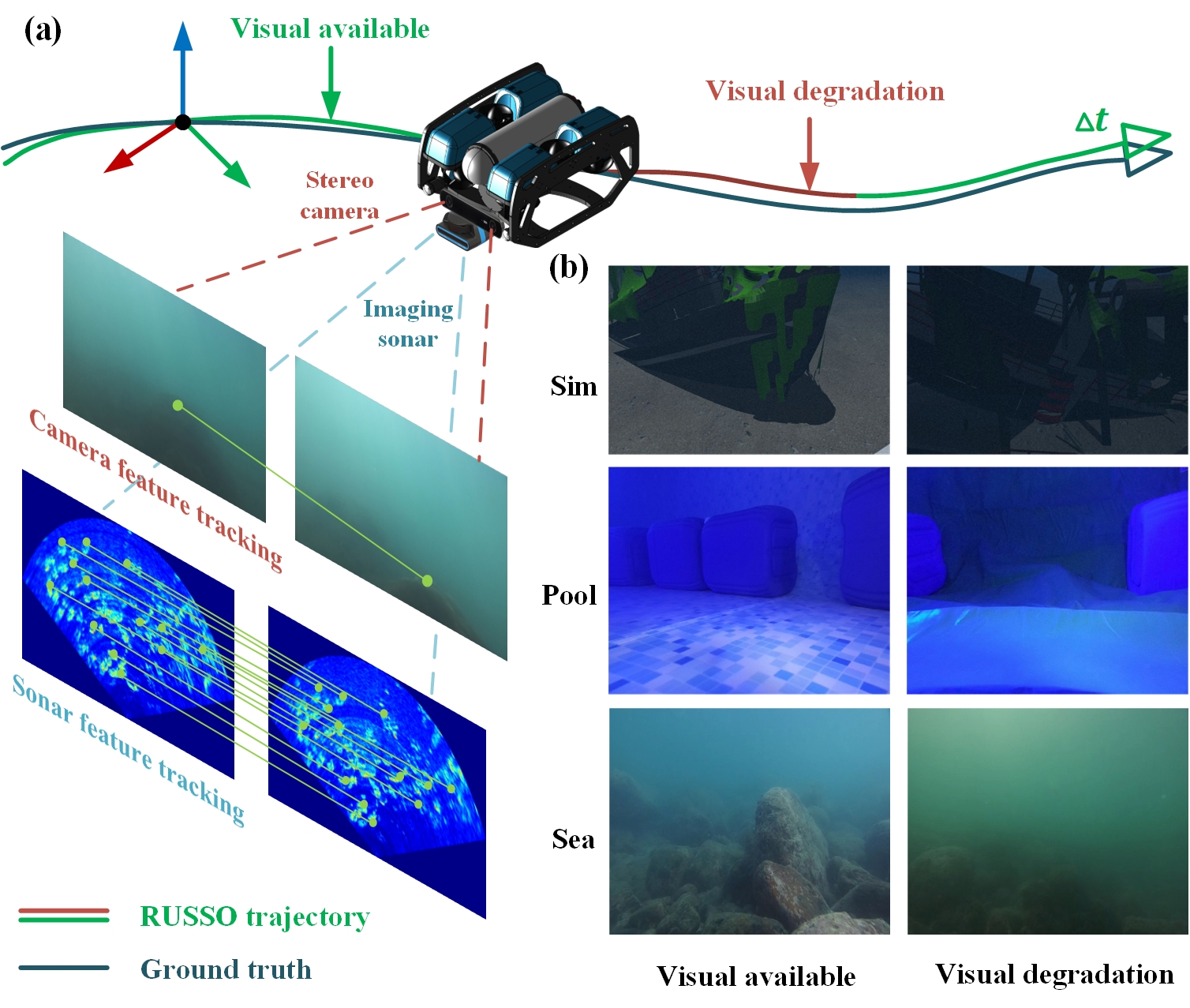}
  \caption{(a) The schematic of our method. The feature tracking of image sonar acts as a constraint to robot pose estimation when visual degradation occurs in $\Delta t$ frame, thus diminishing the pose drift. (b) The samples of visual available and visual degradation scenarios in the underwater simulator, pool, and sea.}
  \label{fig:fig1 image}
\end{figure}

\begin{figure*}[t!]
    \centering
    \includegraphics[width=0.95\linewidth]{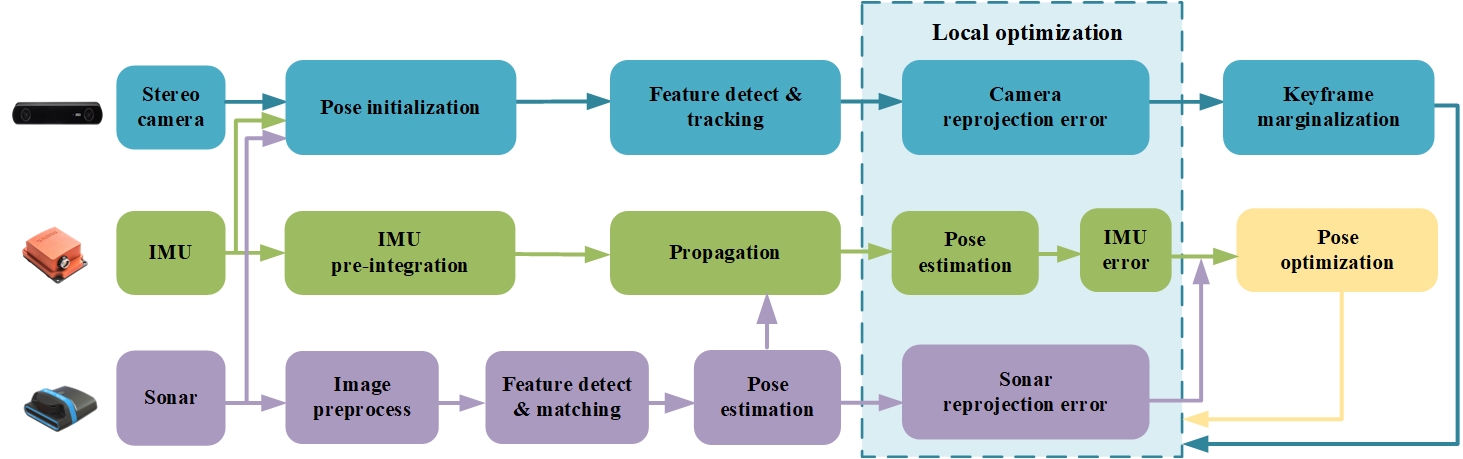}
    \caption{Overview of the RUSSO system, where imaging sonar fusion is integrated with the VIO system. The purple box and lines denote new add-on to the VIO system.}
    \label{fig:algorithm overview}
\end{figure*}

On the other hand, acoustic sensors like sonars are less susceptible to interference than visual cameras, making them ideal for perception in murky underwater environments. To exploit the best of both sensor modalities of sonars and cameras, an underwater SLAM system was developed in \cite{rahman2018sonar} by fusing profiling sonar with a visual-inertial system.  Profiling sonar is well-suited for measuring the depth of ocean floors or other underwater surfaces such as pipelines and caves, but it cannot provide detailed images of underwater scenes, objects, and structures.   Instead, the imaging sonar is good at mapping complex underwater structures.

In this paper, we present a Robust Underwater SLAM with Sonar Optimization (\textbf{RUSSO}) that fuses stereo camera, IMU, and imaging sonar seamlessly for challenging underwater environments. Specifically, we propose to use feature tracking of image sonar as a constraint to pose optimization when suffering from visual degradation, thus diminishing the pose drift. As shown in Fig. \ref{fig:fig1 image}(a). The contributions of our research are four-fold:
\begin{enumerate}
    \item To the best of our knowledge, it is the first underwater SLAM work that fuses imaging sonar with stereo camera and IMU;
    \item A novel IMU propagation optimization method is proposed to provide a good prior using sonar pose estimation during visual degradation, thus enhancing the accuracy of IMU propagation and decreasing localization drift;
    \item To tackle visual initialization failure under challenging environments, a robust SLAM initialization method is proposed to directly utilize the pose estimation between two frames of imaging sonar for initialization;
    \item Extensive experiments, from underwater simulator to real-world lab pool and open sea, are conducted (illustrated in Fig. \ref{fig:fig1 image}(b)), which demonstrate the robustness and accuracy of our proposed RUSSO system in visual degradation environments.
\end{enumerate}

The rest of this paper is organized as follows. Related work is discussed in Section \ref{sec:related}. An overview of our proposed SLAM system RUSSO is followed by an introduction of RUSSO's algorithm details in Section \ref{sec:approach}. Experimental results are shown in Section \ref{sec:experiment}, Finally, conclusions are provided in  Section \ref{sec: conclusion}.

\section{RELATED WORK}\label{sec:related}

Underwater localization has been studied for more than decades, and various methods relying on active sensors and passive sensors have been integrated to build a full SLAM system. Depending on the application scenarios, active acoustic-based positioning methods and passive vision-based localization methods are discussed separately in the following section.

\subsection{Acoustics-based Underwater Positioning}

Traditionally, acoustic-based positioning methods mainly use acoustic sensors like DVL, imaging sonar and side-scan sonar for localization. Proprioceptive sensors like IMU and pressure sensors are fused to improve localization accuracy. Specifically, imaging sonars are employed to provide stable and reliable sensing capability for underwater scenarios. For example,   \cite{johannsson2010imaging} used an imaging sonar to estimate the 3-degree-of-freedom pose for underwater harbour surveillance. An efficient bathymetric SLAM with invalid
loop closure identification using multi-beam sonar was proposed in \cite{9286733}. McConnell et al. \cite{mcconnell2022overhead} proposed to include an overhead image factor by matching sonar images with overhead images. The overhead image factor is similar to the GPS factor in the ground vehicle setting which helps to reduce drift in odometry. To resolve ambiguity in sonar elevation estimation, \cite{mcconnell2020fusing} proposed to combine two orthogonal imaging sonars for sonar mapping. Apart from these, in the field of ground-vehicle application, \cite{burnett2024continuous} proposed an radar-inertial
odometry to resist adverse whether conditions using a spinning mechanical radar, which shares similar sensor
characteristics with imaging sonar. However, all these approaches have limitations in constructing detailed 3D maps which are important for underwater exploration and localization. Therefore, there is an increasing interest in exploring vision-based methods for underwater localization in the marine robotics community.

\subsection{Vision-based Underwater SLAM}
With the development of vision-based SLAM, many vision-only \cite{mur2015orb, wang2017stereo}, as well as visual-inertial methods 
 \cite{campos2021orb, leutenegger2015keyframe, sun2018robust} have been tested in the underwater domain for AUV localization. To address the challenges of visual degradation in the underwater domain, kim et al.
 \cite{kim2013real} proposed to analyze visual saliency to overcome limited field of view and feature-poor regions. More recently, many vision-based SLAM systems fused with other modalities such as IMU, DVL, profiling sonar and depth sensor have been proposed to improve localization accuracy \cite{rahman2018sonar,rahman2019SVIn2,vargas2021robust,xu2021underwater,hu2022tightly}. However, there are still no SLAM methods that fuse imaging sonar with camera and IMU, except very few pioneering works on visual-sonar calibration \cite{raaj20163d, yang2020extrinsic, lindzey2021extrinsic, norman2023actag}.

\section{Method}\label{sec:approach}
In this work, we propose a robust underwater SLAM system RUSSO which combines stereo camera, IMU, and imaging sonar information in a non-linear optimization framework. The system overview of RUSSO is shown in Fig. \ref{fig:algorithm overview}. We propose to fuse imaging sonar based on a visual-inertial odometry (VIO) system and present a novel IMU propagation optimization method which provides a good prior using pose estimation by the imaging sonar when pose estimation deteriorates due to visual degradation. Additionally, we introduce imaging sonar for SLAM initialization if there is a lack of visual features at the initialization stage.

\subsection{Notations}
We define the following notations that are used throughout the paper. As the coordinate frames, $(\cdot)_w$ denotes the world frame, $(\cdot)_b$ denotes the body frame which is defined the same as the IMU frame, $(\cdot)_c$ denotes the camera frame. $s_k$ and $s_i$ are the key and current sonar frame, respectively. $b_k$ and $b_i$ are the key and current body frame, respectively. $\mathbf{T}_{s_ks_i} \in \mathbf{SE}(3)$ denotes pose transformation of the current sonar frame with reference to the sonar keyframe. $\mathbf{P}_{s_i}$ are a set of 3D points in the current sonar frame, and $\mathbf{P}_{s_k}$ are a set of 3D points in the sonar keyframe. Both rotation matrices $\mathbf{R}$ and Hamilton quaternions $\mathbf{q}$ are adopted to represent rotation. $\mathbf{p}_{wb}$, $\mathbf{q}_{wb}$, and $\mathbf{v}_{wb}$ are translation, rotation, and velocity from the body frame to the world frame, respectively.  $\mathbf{T}_{wb} = [\mathbf{q}_{wb}|\mathbf{p}_{wb}] \in \mathbf{SE}(3)$ represents the transformation  from body frame to the world frame, and $\mathbf{g}_{w} = \begin{bmatrix}
    0, 0, g
\end{bmatrix}^{T}$ is the gravity vector in the world frame. $\hat{(\cdot)}$ denotes the noisy measurement or estimate of a certain quantity. $\otimes$ represents the multiplication operation between two quaternions. The robot state is defined as $\mathbf{x}_R= [\mathbf{q}^{T}_{wb},\mathbf{p}^{T}_{wb},\mathbf{v}^{T}_{wb},\mathbf{b}^{T}_g,\mathbf{b}^{T}_a]^T \in SO(3) \times \mathbb{R}^3 \times \mathbb{R}^9$, where $\mathbf{b}_g,\mathbf{b}_a$ represent the gyroscopes and accelerometers bias, respectively. The pose state is defined as $\mathbf{x}_T = [\mathbf{R}^{T}_{wb}, \mathbf{p}^{T}_{wb}]^T$, and the pose error state represented as $\delta\mathbf{x}_T=[\delta\boldsymbol{\theta}^T, \delta\mathbf{p}^T]^T$, where $\delta\boldsymbol{\theta}^T \in \mathbb{R}^3$ is the minimal (3D) axis-angle perturbation for rotation, and can be converted into its quaternion equivalent via exponential mapping.


\begin{figure}[t]
  \centering
  \includegraphics[width=0.48\textwidth]{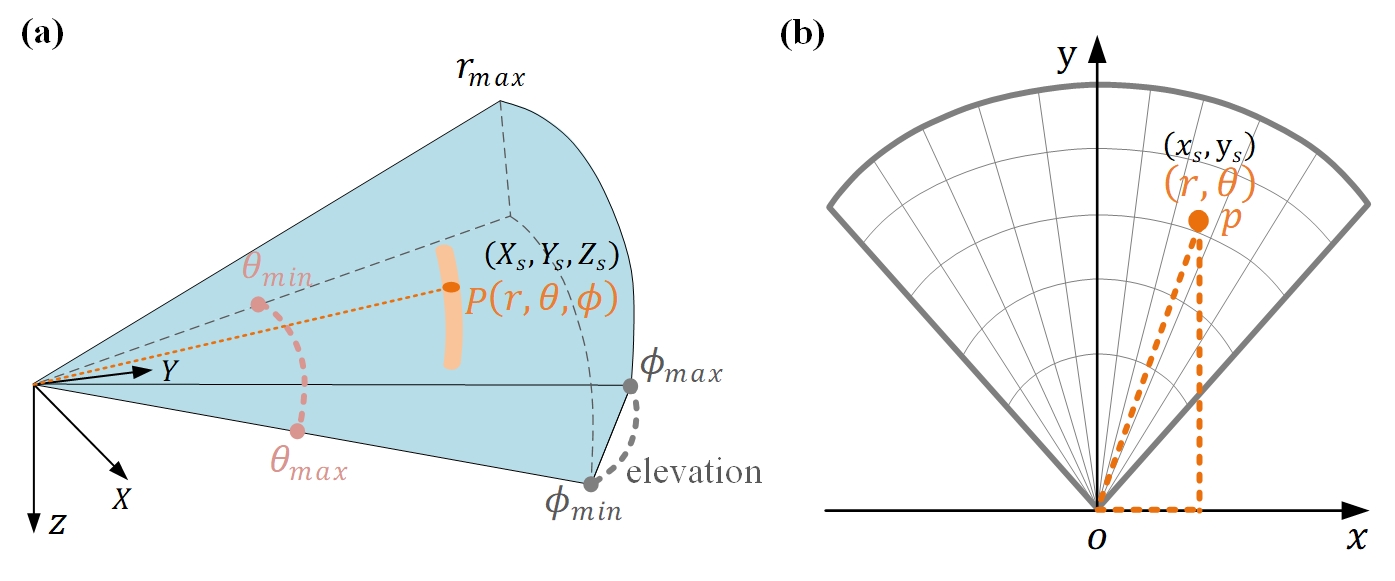}
  \caption{(a) The imaging model of imaging sonar which has FOV in both horizontal and vertical directions. (b) The real sonar image is a 2D image which is a projection of the whole vertical sonar data onto a plane.}
  \label{fig:sonar principle}
\end{figure}

\subsection{Camera Reprojection Error}
The camera reprojection error is formulated as the difference between the camera feature point $\mathbf{z}_c^{i,j,k}$ observed in the image plane and the projection of the corresponding 3D point $\mathbf{P}^{i,j}_c$ on to the image plane, and by using camera projection model $\mathbf{h}_{i}(\cdot)$, the camera reprojection error $\mathbf{e}^{i,j,k}_c$ can be expressed as:
\begin{equation}
\mathbf{e}^{i,j,k}_c = \mathbf{z}_c^{i,j,k} - \mathbf{h}_{i}(\mathbf{P}^{i,j}_{c})
\end{equation}
where $i$ is the camera index, $j$ represents feature point index in camera images observed in the $k^{th}$ frame. Additionally, $\mathbf{P}^{i,j}_{c}$ can be expressed by using landmark $\mathbf{P}^j_w$ in word frame, robot pose $\mathbf{T}_{wb_k}$ in the $k^{th}$ frame, and $\mathbf{T}_{c_ib}=[\mathbf{R}_{c_ib}|\mathbf{p}_{c_ib}] \in \mathbf{SE}(3)$ which donates the calibrated extrinsic parameter from body frame to camera frame, as shown in the following relationship:
\begin{equation}
 \begin{bmatrix}
    x^{i,j,k} \\
    y^{i,j,k} \\
    z^{i,j,k} \\
  \end{bmatrix}=
\mathbf{P}^{i,j}_{c} = \mathbf{R}_{c_ib}\mathbf{R}_{wb_k}^{-1}(\mathbf{P}^j_w-\mathbf{p}_{wb_k})+\mathbf{p}_{c_ib}
\end{equation}
Then, the Jacobian $\mathbf{H}_k$ can be calculated as:
\begin{equation}
    \mathbf{H}_k= \frac{\partial\mathbf{e}^{i,j,k}_c}{\partial\delta\mathbf{x}_T} = \mathbf{H}_{proj}\mathbf{R}_{c_ib}
    \begin{bmatrix}
    \mathbf{H}^{\theta}_k & \mathbf{0}_{3\times9} & \mathbf{H}^{p}_k\\ 
   \end{bmatrix}
\end{equation}
where $\mathbf{H}_{proj},\mathbf{H}^{\theta}_k,\mathbf{H}^{p}_k$ are the Jacobian of the projection $\mathbf{h}_i(\cdot)$ into the $i^{th}$ camera with respect to the landmark in the homogeneous coordinates, orientation, and translation, respectively. They can be expressed as:
\begin{equation}
   \mathbf{H}_{proj}=
    \begin{bmatrix}
    \frac{1}{z^{i,j,k}} & 0 & -\frac{x^{i,j,k}}{z^{i,j,k}}\\
    0 & \frac{1}{z^{i,j,k}} & -\frac{y^{i,j,k}}{z^{i,j,k}}\\
   \end{bmatrix}
\end{equation}
\begin{equation}
  \mathbf{H}^{\theta}_k=
    [\mathbf{R}_{wb_k}^{-1}(\mathbf{P}_{w}^j-\mathbf{p}_{wb_k})]_{\times}
\end{equation}
\begin{equation}
  \mathbf{H}^{p}_k=-\mathbf{R}_{wb_k}^{-1}
\end{equation}
where $[]_{\times}$ corresponds to the skew-symmetric matrix form. 

Given the camera measurement $\mathbf{z}_c^k$, the camera reprojection error term $\mathbf{e}_c^k(\mathbf{x}_T^k, \mathbf{z}_c^k, \mathbf{P}_c)$ is used to correct the robot pose $\mathbf{x}_T^k$. Assuming an approximate normal conditional probability density function $f$ with zero mean and $\mathbf{W}_c^k$ variance $f(\mathbf{e}_c^k|\mathbf{x}_T^k, \mathbf{z}_c^k, \mathbf{P}_c) \approx \mathcal{N}(\mathbf{0},\mathbf{W}_c^k)$, and the conditional covariance $\mathbf{Q}_c(\delta\hat{\mathbf{x}}_T^{k}|\mathbf{z}_c^k,\mathbf{P}_c)$.
then the information matrix $\mathbf{P}_{c}^{k}$ of visual landmarks can be represented as follows:
\begin{equation}
  \mathbf{P}_c^k=\mathbf{W}_c^{k^{-1}}=(\mathbf{H}_k\mathbf{Q}_c(\delta\hat{\mathbf{x}}_T^{k}|\mathbf{z}_c^k,\mathbf{P}_c)\mathbf{H}_k^T)^{-1}
\end{equation}

\subsection{Imaging Sonar Error}
As shown in Fig. \ref{fig:sonar principle}(a), the imaging sonar sensor has different field of view (FOV) in the horizontal and vertical directions. The horizontal FOV is in [$\theta_{min}$, $\theta_{max}$], and the elevation angle is in [$\phi_{min}$, $\phi_{max}$], $r_{max}$ is the maximum detection range of the sonar. Then, a 3D point $\mathbf{P}$ = $(r, \theta, \phi)$ within the sonar FOV in polar coordinate can be represented in Euclidean coordinate as ($X_s$, $Y_s$, $Z_s$) and has the following relationship between the spherical coordinate and Euclidean coordinate 
\begin{equation}
  \begin{bmatrix}
    X_s \\
    Y_s \\
    Z_s \\
  \end{bmatrix}
  =
  \begin{bmatrix}
    r\cos \phi \cos \theta \\
    r\cos \phi \sin \theta \\
    r\sin \phi \\
  \end{bmatrix}
\end{equation}

The sonar image is a projection of the sonar data onto the zero-elevation plane in the vertical direction, therefore losing the elevation angle information. The point $\mathbf{P}$ is now represented as $\mathbf{p} = $ ($r$, $\theta$) in polar coordinate, as shown in Fig. \ref{fig:sonar principle}(b). Each pixel in sonar images may correspond to multiple 3D points. Fox example, the point $\mathbf{p}$ in Fig. \ref{fig:sonar principle}(b) correspond to the orange column in figure (a). We can only get the spatial coordinates of this point in the horizontal plane due to the degradation of sonar elevation angle when converted to the Euclidean coordinate. Therefore, the sonar point $\mathbf{P}_s$ in the 3D space can be denoted as follows:
\begin{equation}
  \mathbf{P}_s =
  \begin{bmatrix}
    x_s \\
    y_s \\
      1 \\
  \end{bmatrix}
  =
  \begin{bmatrix}
    r\cos \theta \\
    r \sin \theta \\
    1 \\
  \end{bmatrix}
\end{equation}
where $r$ denotes the distance of the point from sonar, and has the following relationship
\begin{equation}
r = \sqrt{(u_s -W_s/2)^2 + (H_s - v_s)^2} * r_{resolusion}
\end{equation}
and $W_s$ and $H_s$ are the width and height of the sonar image, respectively. ($u_s, v_s$) is the pixel coordinate of the sonar point while $r_{resolusion}$ represents the pixel resolution of sonar image, which can be calculated with:
\begin{equation}
r_{resolusion} = r_{max} / H_s
\end{equation}
\begin{figure}[ht]
  \centering
  \includegraphics[width=0.48\textwidth]{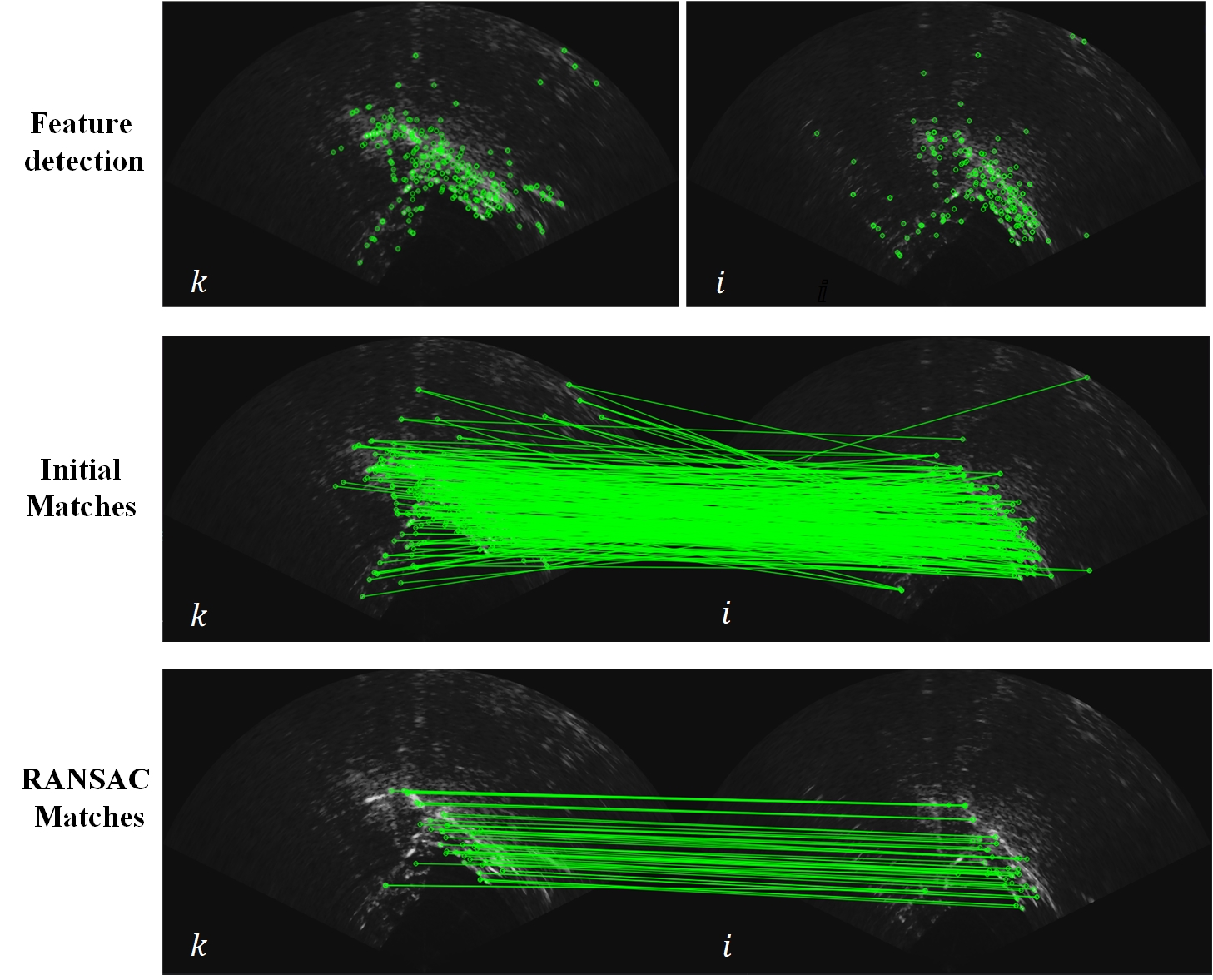}
  \caption{The green circles represent the extracted feature points of sonar image using A-KAZE. The initial matching contains many incorrect matches. Outliers are rejected after using RANSAC algorithm, and consequently feature points are correctly matched between the keyframe and the current frame. }
  \label{fig:sonar features}
\end{figure}

Since sonar images have relatively low signal-to-noise ratio plus speckled noise, they are first preprocessed to filter out the noise using the media filter. After that, A-KAZE \cite{alcantarilla2013fast} is used to detect and match features. A-KAZE has been proved to be consistent in feature detection and matching in nonlinear scale space, while the use of median filter can preserve texture in nonlinear scale space \cite{shin2015bundle}.  We first extract A-KAZE feature points from the sonar images. However, in practice there exist incorrect matches after the initial feature matching. Here, we adopt 2D-2D random sample consensus (RANSAC) in the 2D space of sonar points to refine correspondences, which is based on the fundamental matrix estimation. The filtered result is shown in Fig. \ref{fig:sonar features}.

After receiving reasonable associations, the two sets of feature points $\mathbf{P}_{s_k}$ and $\mathbf{P}_{s_i}$, between keyframe and current frame are used to compute the relative pose $\mathbf{T}_{s_ks_i} = 
  \begin{bmatrix}
    \mathbf{R}_{2\times2} &  \mathbf{0}_{2\times1} & \mathbf{p}_{2\times1} \\
    \mathbf{0}_{1\times2} & 1 & 0 \\
    \mathbf{0}_{1\times2} & 0 & 1 \\
  \end{bmatrix} \in SE(3)$, which is transformed into a 4 $\times$ 4 homogeneous matrix to ensure the consistency in the matrix form of pose. In this work, the $\mathbf{T}_{s_ks_i}$ does not account for roll and pitch motion, as these are not observable by a 2D imaging sonar.
\begin{equation}
\mathbf{P}_{s_k} = \mathbf{T}_{s_ks_i} \mathbf{P}_{s_i}
\label{eq:sonar}
\end{equation}

  
Assuming $\mathbf{P}_{s_k}$ and $\mathbf{P}_{s_i}$ are associated sonar landmarks observed in $k^{th}$ and $i^{th}$ sonar frame, and $\mathbf{T}_{wb_{k}}$ and $\mathbf{T}_{wb_i}$ are the correspondent robot pose in world frame, then we can formulate the sonar reprojection error, as below:
\begin{equation}
\mathbf{e}_{s} = \mathbf{T}_{wb_i}\mathbf{T}_{bs}\mathbf{P}_{s_{i}} - \mathbf{T}_{wb_{k}}\mathbf{T}_{bs}\mathbf{P}_{s_{k}}
\end{equation}
where $\mathbf{T}_{bs} \in \mathbf{SE}(3)$ denotes the transformation matrix from sonar coordinate frame to IMU coordinate frame,  which is obtained via extrinsic calibration.

To optimize the keyframe pose $\mathbf{T}_{wb_k}$, the Jacobian matrix $\mathbf{H}_s$ can be derived as:
\begin{equation}
\mathbf{H}_s = \frac{\partial\mathbf{e}_{s}}{\partial\delta\mathbf{x}_T}=
 \begin{bmatrix}
    -[\mathbf{T}_{wb_k}\mathbf{T}_{bs}\mathbf{P}_{s_k}]_{\times} & \mathbf{I}_{3\times3}
 \end{bmatrix}
\end{equation}

Given the sonar measurement $\mathbf{z}_s^k$, $\mathbf{z}_s^i$ and the error term $\mathbf{e}_s^k(\mathbf{x}_T^k, \mathbf{x}_T^i, \mathbf{z}_s^k, \mathbf{z}_s^i)$ of imaging sonar at frame $k^{th}$ and $i^{th}$ can be used to optimize the robot state $\mathbf{x}_T^k$. We assume an approximate normal conditional probability density function $f$ with zero mean and $\mathbf{W}_s^k$ variance $f(\mathbf{e}_s^k|\mathbf{x}_T^k, \mathbf{x}_T^i, \mathbf{z}_s^k, \mathbf{z}_s^i) \approx \mathcal{N}(\mathbf{0},\mathbf{W}_s^k)$, and the conditional covariance $\mathbf{Q}(\delta\hat{\mathbf{x}}_T^{k}|\mathbf{x}_T^i, \mathbf{z}_s^k, \mathbf{z}_s^i)$.
Then the information matrix of sonar landmark is:
\begin{equation}
\mathbf{P}_s^k=\mathbf{W}_s^{k^{-1}} =(\mathbf{H}_s
\mathbf{Q}(\delta\hat{\mathbf{x}}_T^{k}|\mathbf{x}_T^i, \mathbf{z}_s^k, \mathbf{z}_s^i)
\mathbf{H}_s^T
)^{-1}
\end{equation}
\subsection{IMU Error}
The raw gyroscope and accelerometer measurements $\hat{\boldsymbol{\omega}}_{b_t}$, $\hat{\mathbf{a}}_{b_t}$ is modelled by:
\begin{equation}
\begin{aligned}
  \hat{\boldsymbol{\omega}}_{b_t} &=  \boldsymbol{\omega}_{b_t} + \mathbf{b}_{g} +  \mathbf{n}_{g}\\
  \hat{\mathbf{a}}_{b_t} &=  \mathbf{a}_{b_t} + \mathbf{q}_{b_tw}\mathbf{g}_{w} + \mathbf{b}_{a} +  \mathbf{n}_{a}\\
\end{aligned}
\end{equation}
The measurements are measured in the body frame and affected by acceleration bias $\mathbf{b}_{a}$, gyroscope
bias $\mathbf{b}_{g}$, acceleration noise $\mathbf{n}_{a} \sim \mathcal{N}(0, \boldsymbol{\sigma}_a^2)$ and gyroscope noise $\mathbf{n}_{g}\sim \mathcal{N}(0, \boldsymbol{\sigma}_g^2)$, where the acceleration and gyroscope bias are modeled as random walk, whose derivatives are Gaussian white noise, $\mathbf{n}_{b_a} \sim \mathcal{N}(0, \boldsymbol{\sigma}_{b_g}^2)$, $\mathbf{n}_{b_a} \sim \mathcal{N}(0, \boldsymbol{\sigma}_{b_g}^2)$. $\mathbf{q}_{b_tw}$ is rotation from the world frame to the body frame.
  
In the IMU propagation step, the IMU pre-integration terms are calculated as below:
\begin{equation}\label{pre-integration}
\begin{aligned}
\boldsymbol{\alpha}_{b_{i}b_{i+1}} = \int\int_{t\in[i, {i+1}]}\mathbf{q}_{b_ib_t}(\mathbf{a}_{b_t}-\mathbf{b}_{a_t})\delta t^2 \\ 
\boldsymbol{\beta}_{b_{i}b_{i+1}} = \int_{t\in[i, {i+1}]}\mathbf{q}_{b_ib_t}(\mathbf{a}_{b_t}-\mathbf{b}_{a_t})\delta t \\
\mathbf{q}_{b_{i}b_{i+1}} = \int_{t\in[i, {i+1}]}\mathbf{q}_{b_ib_t}\otimes
\begin{bmatrix}
    0 \\
    \frac{1}{2}\boldsymbol{(\omega}_{b_t}- \mathbf{b}_{g_t}) \\
  \end{bmatrix}\delta t
\end{aligned}
\end{equation}
where rotation $\mathbf{q}_{b_ib_t}$ denotes the integration of the IMU data at frame $b_t$ with respect to the reference frame $b_i$. $\boldsymbol{\alpha}_{b_{i}b_{i+1}}$, $\boldsymbol{\beta}_{b_{i}b_{i+1}}$, and $\mathbf{q}_{b_{i}b_{i+1}}$ are only related to IMU biases instead of other states between $b_i$ and $b_{i+1}$. More detailed derivation of IMU pre-integration can be referred to \cite{lupton2011visual}.

We can derive continuous-time linearized model of error term of (\ref{pre-integration}) as follows:
\begin{equation}
\begin{split}
\begin{bmatrix}
    \dot{\boldsymbol{\alpha}}_{b_{i}t} \\
    \dot{\boldsymbol{\beta}}_{b_{i}t} \\
    \dot{\boldsymbol{\theta}}_{b_{i}t} \\
    \dot{\mathbf{b}}_{a_t} \\
    \dot{\mathbf{b}}_{g_t}\\
\end{bmatrix} = 
\begin{bmatrix}
    \mathbf{0} & \mathbf{I} & \mathbf{0} & \mathbf{0} & \mathbf{0}  \\
    \mathbf{0} & \mathbf{0} & -\mathbf{q}_{b_{i}t}[\mathbf{a}_{b_t}-\mathbf{b}_{a_t}]_\times & -\mathbf{q}_{b_{i}t} & \mathbf{0}  \\
    \mathbf{0} & \mathbf{0} & -[\boldsymbol{\omega}_{b_t}- \mathbf{b}_{g_t}]_\times  & \mathbf{0} & -\mathbf{I}\\
    \mathbf{0} & \mathbf{0} & \mathbf{0} & \mathbf{0} & \mathbf{0} \\
    \mathbf{0} & \mathbf{0} & \mathbf{0} & \mathbf{0} & \mathbf{0} \\
\end{bmatrix}
\begin{bmatrix}
    \boldsymbol{\alpha}_{b_{i}t} \\
    \boldsymbol{\beta}_{b_{i}t} \\
    \boldsymbol{\theta}_{b_{i}t} \\
    \mathbf{b}_{a_t} \\
    \mathbf{b}_{g_t}\\
\end{bmatrix}\\
+ \begin{bmatrix}
    \mathbf{0} & \mathbf{0} & \mathbf{0} & \mathbf{0}  \\
    -\mathbf{q}_{b_{i}t} & \mathbf{0} & \mathbf{0} & \mathbf{0} \\
    \mathbf{0} & -\mathbf{I} & \mathbf{0} & \mathbf{0}\\
    \mathbf{0} & \mathbf{0} & \mathbf{I} & \mathbf{0} \\
    \mathbf{0} & \mathbf{0} & \mathbf{0} & \mathbf{I}\\
\end{bmatrix}
\begin{bmatrix}
    \mathbf{n}_a \\
    \mathbf{n}_g \\
    \mathbf{n}_{b_a} \\
    \mathbf{n}_{b_g} \\
\end{bmatrix} = \mathbf{F}_t\delta\mathbf{z}_{b_it} + \mathbf{G}_t\mathbf{n}_t
\end{split}
\end{equation}
The covariance propagation equation can be computed recursively by a first-order discrete-time covariance update, that is, the convariance $\mathbf{P}_I^{b_{i+1}}$ can be updated from the convariance $\mathbf{P}_I^{b_{i}}$ for the $i^{th}$ IMU measurement, taking the following form:
\begin{equation}
    \mathbf{P}_I^{b_{i+1}} = (\mathbf{I} + \mathbf{F}_t\delta t)\mathbf{P}_I^{b_{i}}(\mathbf{I} + \mathbf{F}_t\delta t)^T + \delta t \mathbf{G}_t\mathbf{Q}_t\mathbf{G}_t^T
\end{equation}
where $t\in [i,i+1]$, and $\delta t$ is the time between two IMU measurements, $\mathbf{Q}_t =$ diag$(\boldsymbol{\sigma}_a^2,\boldsymbol{\sigma}_g^2,\boldsymbol{\sigma}_{b_a}^2,\boldsymbol{\sigma}_{b_g}^2)$ is the continous-time noise covariance matrix.

Given the IMU pre-integration terms, the position $\mathbf{p}_{wb_{i+1}}$, velocity $\mathbf{v}_{wb_{i+1}}$, and rotation $\mathbf{q}_{wb_{i+1}}$ in frame $b_{i+1}$ can be expressed by:
\begin{equation}
\begin{aligned}
  \mathbf{p}_{wb_{i+1}} &=    \mathbf{p}_{wb_i} + \mathbf{v}_{wb_i}\Delta t - \frac{1}{2}\mathbf{g}_w\Delta t^2 + \mathbf{q}_{wb_i}\boldsymbol{\alpha}_{b_{i}b_{i+1}}\\
  \mathbf{v}_{wb_{i+1}} &= \mathbf{v}_{wb_i} - \mathbf{g}_w \Delta t + \mathbf{q}_{wb_i}\boldsymbol{\beta}_{b_{i}b_{i+1}}\\
  \mathbf{q}_{wb_{i+1}} &=    \mathbf{q}_{wb_i}\mathbf{q}_{b_{i}b_{i+1}} \\
\end{aligned}
\end{equation}
So, it is obvious that the state of IMU propogation in frame $b_{i+1}$ is determined by the state in frame $b_{i}$, since the state $\mathbf{T}_{wi}$ in frame $b_{i}$ contains position $\mathbf{p}_{wb_i}$ and rotation $\mathbf{q}_{wb_i}$.


However, the VIO system loses the observation from camera when suffer from visual degradation, which means the current state will deteriorate, and the pose estimation that relies on the IMU propagation will drift quickly due to IMU noise. Here we try to improve the current state during visual degradation in order to reduce drift.

To provide the IMU propagation process with a better prior, we propose using the estimation $\mathbf{T}_{s_ks_i}$ from imaging sonar to improve the current state $\mathbf{T}_{wi}$. By substituting  $\mathbf{T}_{wi}$ with $\hat{\mathbf{T}}_{wi}$, $\hat{\mathbf{T}}_{wi}$ is computed as:
\begin{equation}
\hat{\mathbf{T}}_{wi} = \mathbf{T}_{wb_{k}}\mathbf{T}_{bs}\mathbf{T}_{s_ks_i}
\end{equation}
Now we decompose $\hat{\mathbf{p}}_{wb_i}$ and $\hat{\mathbf{q}}_{wb_i}$ from $\hat{\mathbf{T}}_{wi}$. The IMU error term  $\mathbf{e}_{I} = [\mathbf{e}_{I}^p, \mathbf{e}_{I}^v, \mathbf{e}_{I}^q, \mathbf{e}_{I}^{ba}, \mathbf{e}_{I}^{bg}]$ is expressed as follows:
\begin{equation}
\begin{aligned}
  \mathbf{e}_{I}^p  &=  \hat{\mathbf{q}}_{b_{i}w}(\mathbf{p}_{wb_{i+1}} - \hat{\mathbf{p}}_{wb_i} - \mathbf{v}_{wb_i}\Delta t + \frac{1}{2}\mathbf{g}_w\Delta t^2) - \boldsymbol{\alpha}_{b_{i}b_{i+1}}\\
  \mathbf{e}_{I}^v &= \hat{\mathbf{q}}_{b_{i}w}(\mathbf{v}_{wb_{i+1}} - \mathbf{v}_{wb_i} + \mathbf{g}_w \Delta t) -\boldsymbol{\beta}_{b_{i}b_{i+1}} \\
  \mathbf{e}_{I}^q &= 2[(\mathbf{q}_{b_{i}b_{i+1}})^{-1} \otimes(\hat{\mathbf{q}}_{b_{i}w}\otimes\mathbf{q}_{wb_{i+1}})]_{xyz} \\
  \mathbf{e}_{I}^{ba} &=  \mathbf{b}_{a_{i+1}} - \mathbf{b}_{a_i} \\
  \mathbf{e}_{I}^{bg} &=  \mathbf{b}_{g_{i+1}} - \mathbf{b}_{g_i}
\end{aligned}
\end{equation}
where $[\cdot]_{xyz}$ means taking the vector part from a quaternion. 


Finally, we construct the cost function $\mathbf{J}(\mathbf{X})$, which is the sum of three different terms, that is the camera reprojection error $\mathbf{e}_c$, the IMU error $\mathbf{e}_I$ and imaging sonar reprojection error $\mathbf{e}_s$:
\begin{equation}
\begin{split}
\mathbf{J}(\mathbf{X}) = &\alpha\sum_{k=1}^{K}\sum_{j\in k}\mathbf{e}_{s}^{j,k^T}\mathbf{P}_{s}^{k}\mathbf{e}_{s}^{j,k}  +\sum_{k=1}^{K-1}\mathbf{e}_{I}^{k^T}\mathbf{P}_{I}^{k}\mathbf{e}_{I}^{k} \\ &+\sum_{i=1}^{I=2}\sum_{k=1}^{K}\sum_{j\in\gamma(i,k)} \mathbf{e}_{c}^{i,j,k^T}\mathbf{P}_{c}^{k}\mathbf{e}_{c}^{i,j,k}\end{split}
\end{equation}
where $i$ is the camera index, $j$ represents feature point index in camera or sonar images observed in the $k^{th}$ frame. $\mathbf{P}_{c}^{k}$, $\mathbf{P}_{I}^{k}$ and $\mathbf{P}_{s}^{k}$ are the information matrix of visual landmarks, IMU, sonar landmarks, respectively. The value of $\alpha$ is adapted empirically to adjust the weight of residual of image sonar, making the local optimization pay more attention to the sonar observations during visual degradation (empirically when camera matched feature points falls below 10). In this way, the imaging sonar residual is used as a strong constraint to diminish the errors introduced by camera degradation.

\begin{algorithm}
  \SetAlgoLined
  \textbf{Input:} stereo camera image $C \in \mathbb{R}^{m \times n} $\ , sonar image $S \in \mathbb{R}^{m \times n} $\;
  \textbf{Parameters:} Minimum correspondences number of camera feature points $\delta_{c}$, estimated sonar pose $\mathbf{T}_{s_ks_i}$ = [$\mathbf{R}_{s_ks_i}$$|$$\mathbf{t}_{s_ks_i}$]\;
  Set the first camera image as keyframe $C_k$\;
  Set the first sonar image as keyframe $S_k$\;
  $init \gets False$ 
  
  \While{$init \neq True$}{
  
      Get number of feature matches $n$ between keyframe $C_k$ and current frame $C_i$\;
      \If {n $\leq  \delta_{c}$}
      {
         Extract feature points $\mathbf{P}_{s_k}$ and $\mathbf{P}_{s_i}$\;
         Estimate the relative sonar pose $\mathbf{T}_{s_ks_i}$\;
         \If {$\mathbf{R}_{s_ks_i} \neq \mathbf{I}$ or $\mathbf{t}_{s_ks_i} \neq \mathbf{0}$}
         {
           $init \gets$ True\;
         }
      }
  }
  \caption{SLAM initialization algorithm}
  \label{algo:initialization}
\end{algorithm}

\begin{table*}
    \footnotesize
    \centering
    \caption{Sequence Length of experimental Dataset} 
    \label{tab:sequence_length}
    \scalebox{1.1}{
    \begin{tabular}{ l|p{1.4cm}|p{1.4cm}|p{1.4cm}|p{1.4cm}|p{2.5cm}|p{1.4cm}|p{1.4cm}}
    \hline
    Sequence    & Simulator & Pool 1 & Pool 2 & Pool 3 & Initialization scenario & Sea 1 & Sea 2 \\ \hline
    Length (m) & 86.73 & 18.52 & 18.62 & 14.93 & 12.46 & 58.12 & 60.35   \\ \hline
    \end{tabular}
    }
\end{table*}

\begin{table*}
    \footnotesize
    \centering
    \caption{RMSE (m) of the localization in all sequences}
    \label{tab:RMSE}
    \begin{tabular}{ l|p{1.7cm}p{1.6cm}p{1.6cm}p{1.6cm}p{2.7cm}p{1.6cm}p{1.6cm}}
    \hline
    & \multicolumn{7}{c}{\textbf{Sequence}} \\
    \textbf{Method} &Simulator & Pool 1 & Pool 2 & Pool 3 & Initialization scenario  & Sea 1 & Sea 2 \\
    \hline
    RUSSO        &\textbf{0.898}  & \textbf{0.219}  & \textbf{0.277} & \textbf{0.194} & \textbf{0.102} & \textbf{1.409} & \textbf{1.884}\\
    \hline
    SVIn2 (VI)   & 2.108 & 0.351  & 0.329 & 0.396 & N/A  & 3.667 & 4.028\\
    \hline
    VINS-Fusion   &N/A &0.619 &0.391 &0.292 &0.401 & 2.354 & N/A\\
    \hline
    \end{tabular}\par
    \smallskip
    The N/A results indicate that the SLAM fails to initialize or loses tracking in this sequence. 
\end{table*}

\begin{figure}[h!]
  \centering
  \includegraphics[width=0.48\textwidth]{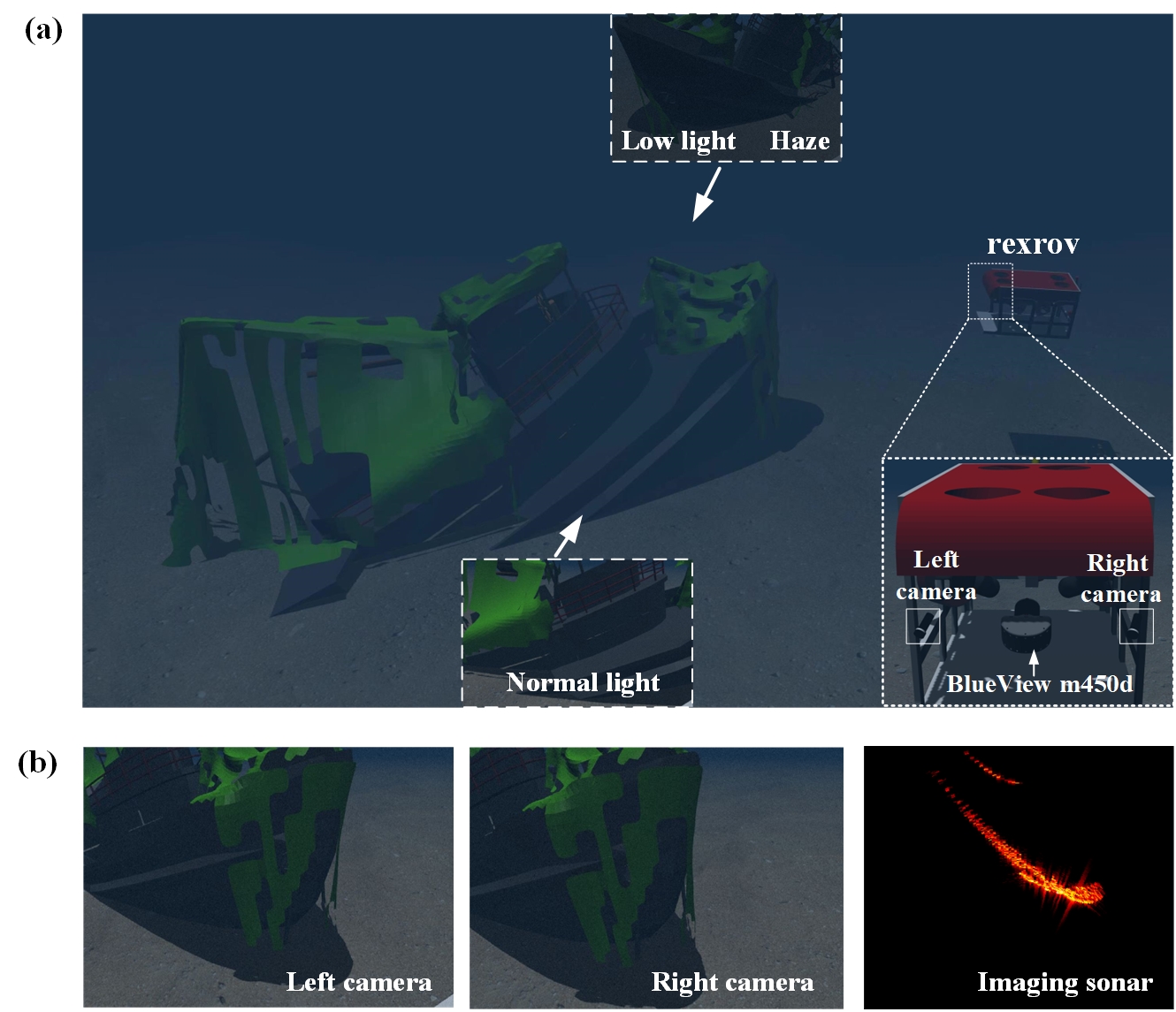}
  \caption{(a) The experimental environment with varying light and haze in UUV simulator. (b) The stereo camera image and corresponding imaging sonar image of rexrov robot.}
  \label{fig:simulator environment}
\end{figure}

\begin{figure}[h]
  \centering
  \includegraphics[width=0.48\textwidth]{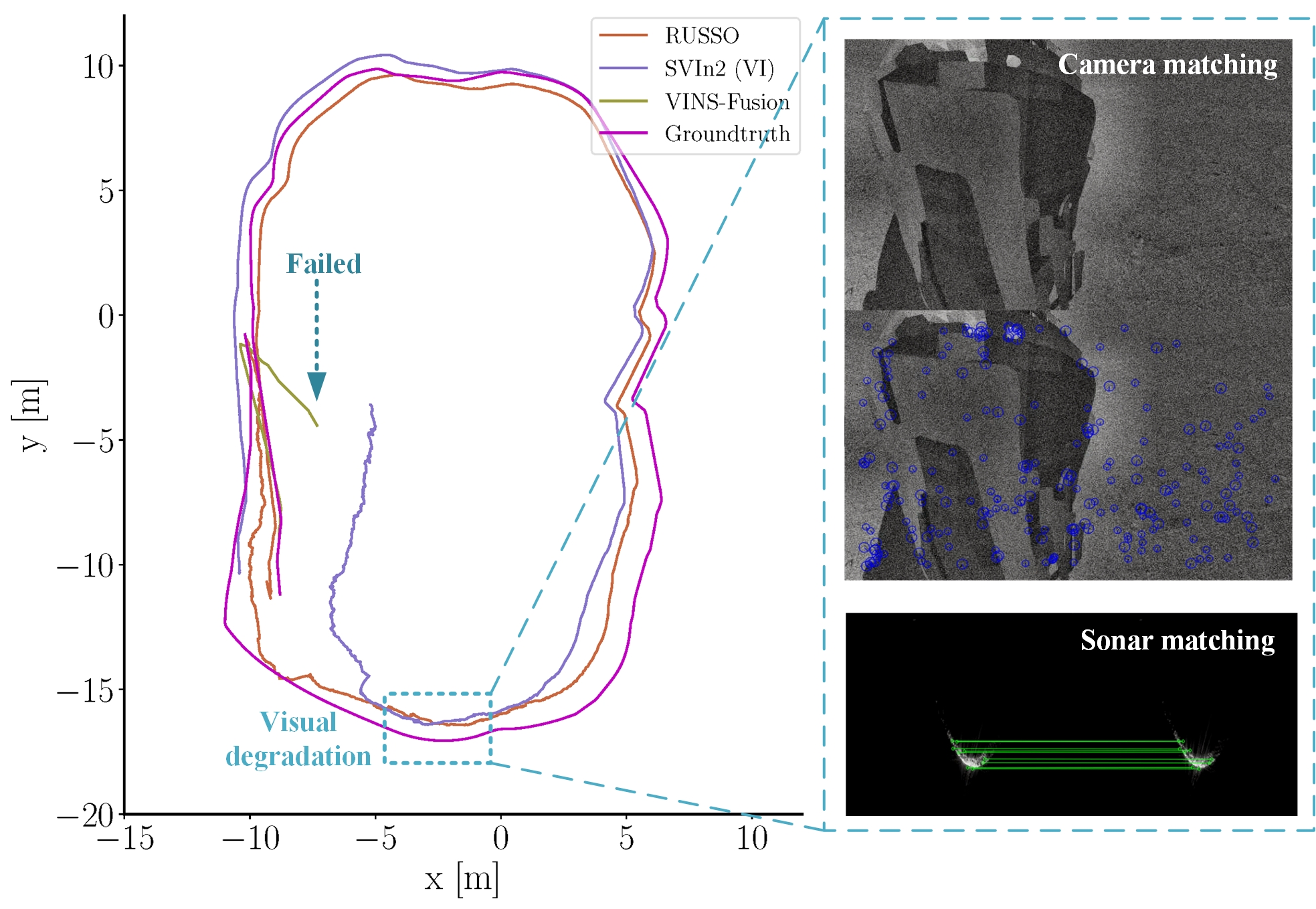}
  \caption{The experimental result in underwater simulator. The blue dotted box shows that the trajectory of SVIn2 drift rapidly when camera feature matching failed, but our method still remain stable and has high performance. The blue dotted arrow shows VINS-Fusion failed. }
  \label{fig:simulator trajectory}
\end{figure}
\subsection{SLAM Initialization}
Tightly coupled SLAM systems are highly nonlinear and require accurate initialization for successful state estimation. They heavily rely on the accuracy of initial values, including gravity, velocity, bias, and feature depth. Poor initialization can reduce convergence speed or even result in completely incorrect estimates.
The vision-based or VIO SLAM systems require sufficient visual feature points during the initialization stage. However, in underwater environment such as in the open sea, visual degradation is commonly encountered. There might be insufficient number of features within the visable range and thus initialization might fail. Imaging sonars can perceive of environment at an extended range and detect points directly from the scene within typically 10 to a few hundreds meters. To maintain a stable initialization, we propose to use pose estimation from the imaging sonar and IMU in visual degradation scenarios.

The number of visual feature correspondences is used to evaluate the healthiness of the visual initialization. If the number of matched features is below a threshold $\delta_{c}$, we will skip the pose estimation from vision and use sonar and IMU for initialization.  The initialization process is considered successful when both the translation and rotation estimation works. The detail is summarized in Algorithm \ref{algo:initialization}.

\section{EXPERIMENTAL RESULTS}\label{sec:experiment}
To adequately validate the feasibility of RUSSO system, we have performed extensive experimental evaluations in visually challenging scenarios in the underwater simulator, lab pool, and open sea. The sequence lengths of all scenarios are shown in Table \ref{tab:sequence_length}. In all experiments, the state-of-the-art underwater SLAM algorithm SVIn2 \cite{rahman2019SVIn2} and VIO algorithm VINS-Fusion \cite{qin2017vins} are used for comparison. Since the underwater simulator platform cannot simulate a profiling sonar and we don't have a profiling sonar for real experiments as well, the sonar option of SVIn2 is disabled and only the camera and IMU sensor are used and hence denoted as SVIn2 (VI). Additionally, loop closures are disabled in all experiments for fair comparison. What's more, the same stereo camera and IMU data are used in each comparison experiment.

\begin{figure*}[t!]
    \centering
    \includegraphics[width=1\linewidth]{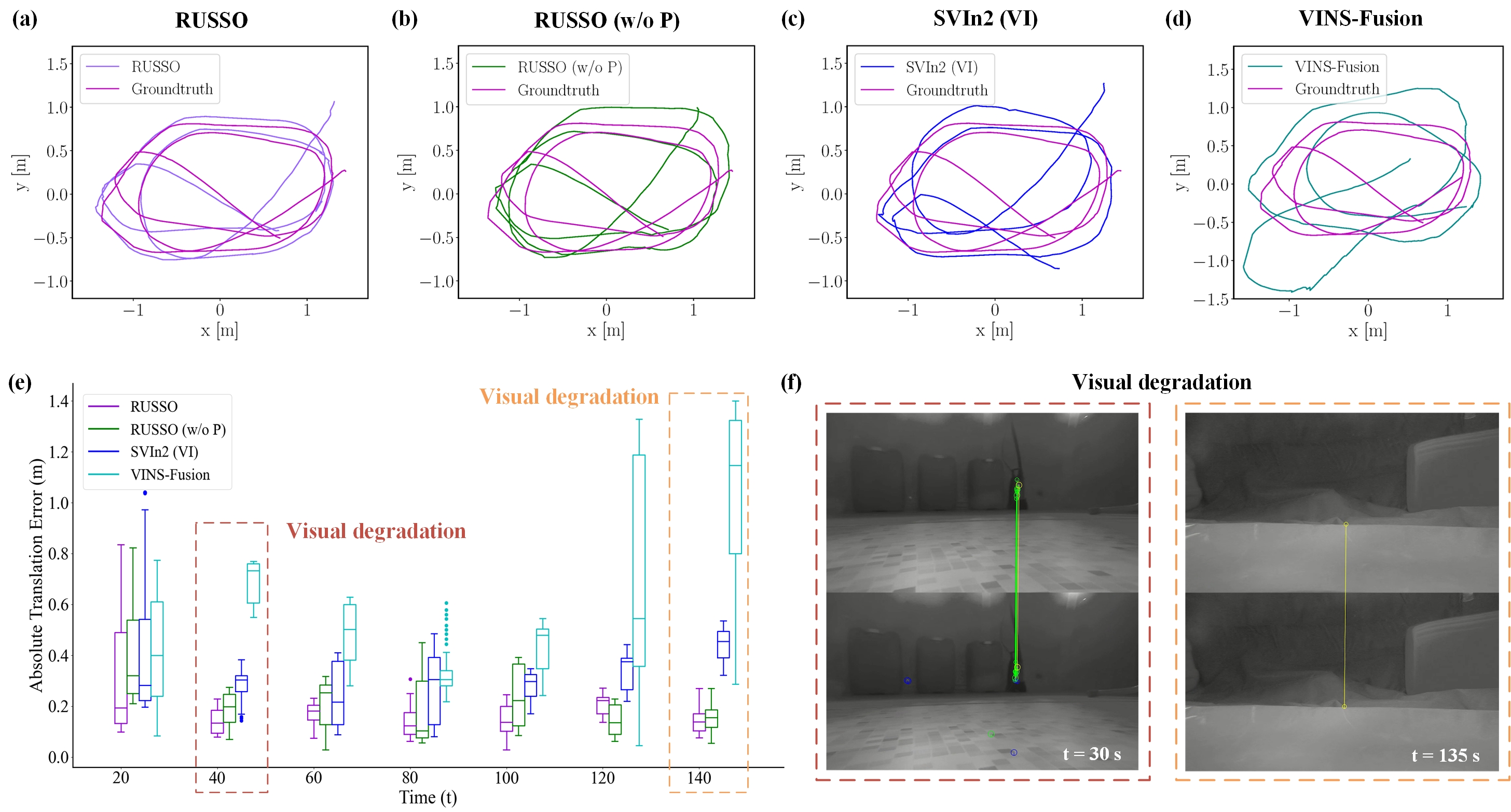}
    \caption{Results of the pool 1 sequence. (a) The localization result of our proposed SLAM system. (b) The localization result of our proposed method without sonar prior for IMU propagation, "w/o P" means without sonar prior. (c) The localization result of SVIn2 with sonar option disabled, "VI" indicates the use of visual and IMU. The localization result on VIO system of SVIn2 which was disabled sonar option, "VI" indicates the use of visual and IMU. (d) The localization result using VINS-Fusion. (e) Comparison result of absolute translation error per 20 s for all algorithms, the red pink and yellow dotted box represent robot suffer from visual degradation within the sample time. (f) Samples of camera feature matching when visual degradation occur, there are few feature points that can be matched.}
    \label{fig:pool trajectory}
\end{figure*}

\subsection{Simulation Experiments}
We first test our proposed RUSSO in a simulation environment. A BlueView m450d imaging sonar simulated by DAVE simulator \cite{zhang2022dave} is integrated into a rexrov robot with a stereo camera and IMU sensors in UUV simulator \cite{manhaes2016uuv}. The robot operating system (ROS) is used to control the robot and collect sensor data from the simulator platform. The stereo camera and sonar image in the corresponding position are shown in Fig. \ref{fig:simulator environment}(b). Then, we place the rexrov robot into a shipwreck simulation scenario with varying light and haze intensities, intentionally designed as visually challenging conditions, as shown in Fig. \ref{fig:simulator environment}(a). The robot is driven around the shipwreck with sensor data collected for algorithm validation. In the simulator experiments, the contrast limited adaptive histogram equalization (CLAHE) filter \cite{pisano1998contrast} is used to enhance image contrast in the prepossessing while reducing the noise amplification.

\begin{figure}[h]
  \centering
  \includegraphics[width=0.48\textwidth]{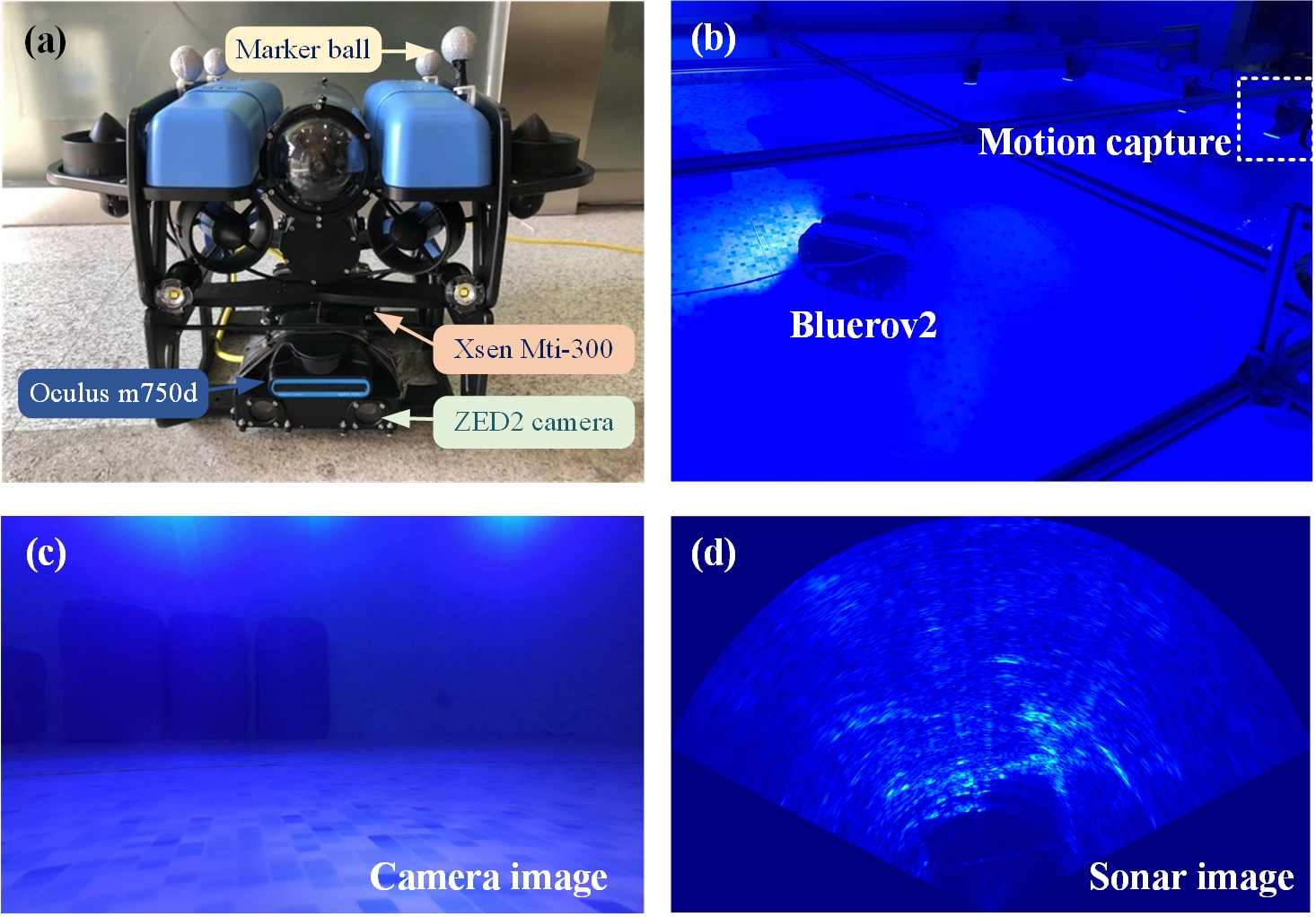}
  \caption{(a) The experimental platform of ours. (b) The experiment environment in pool setting up with motion capture system. (c) The ZED2 camera image in the pool. (d) The Oculus m750d image in the pool.}
  \label{fig:pool environment}
\end{figure}

As shown in Fig. \ref{fig:simulator trajectory}, both SVIn2 (VI) and VINS-Fusion perform well under good visual conditions, but as the
visual condition degrades, VINS-Fusion fails quickly, and the drift of SVIn2 (VI) increases significantly. In contrast, with imaging sonar constraint, our proposed RUSSO remains robust and accurate all the time. The quantitative results of estimation errors in Table \ref{tab:RMSE} further justify the above conclusion.

\subsection{Experiments in The Lab Pool}
Apart from the simulator experiments, we also conduct real-world experiments in a lab pool. The pool is about 3m in width and 7m in length. A motion capture system NOKOV with 12 underwater cameras mounted in the pool is used to provide the ground truth poses of the underwater robot, as shown in Fig. \ref{fig:pool environment}(b).  We use a Bluerov2 robot with eight-thruster configurations as the robot platform. The robot is equipped with a ZED2 stereo camera, a high-precision Xsens MTi-300 IMU, and Oculus m750d imaging sonar, as shown in Fig. \ref{fig:pool environment}(a). Sample camera and sonar images obtained in the pool are shown in \ref{fig:pool environment}(c)-(d). For more accurate intrinsic and extrinsic parameters, all three different sensors are well-calibrated underwater before experiments. Due to the relatively short trajectories in the pool experiments, we use the low-cost IMU embedded in the ZED2 stereo camera rather than the high-precision Xsens IMU, which better demonstrates the performance differences among the three SLAM algorithms for comparisons. 

We collect two sequences in the pool environment. Sequence 1 contains two visual degradation periods between 30-45 seconds and 130-145 seconds while sequences 2 and 3 do not suffer from visual degradation. To manually create visual degradation, we cover the wall and bottom of half of the pool with a pure black plastic sheet and the other half part uncovered. As shown in Fig. \ref{fig:fig1 image}(b), almost no visual feature points can be extracted from the covered part of the pool.

\subsubsection{Algorithm Comparison} All three algorithms have been run in both sequences and the RMSE of localization error is shown in Table \ref{tab:RMSE}. RUSSO has higher localization accuracy in both sequences and outperforms other algorithms sharply in sequence pool 1 with visual degradation. Specifically, the estimated and ground-truth trajectories in sequence pool 1 are plotted in Fig. \ref{fig:pool trajectory}(a), (c), (d) for qualitative comparison. Fig. \ref{fig:pool trajectory}(e) plots the absolute translation errors over each 20 seconds, which highlights localization performance impacted by visual degradation of all comparative SLAM algorithms. It verifies that our proposed RUSSO remains robust against visual degradation. 

\begin{figure}[h]
  \centering
  \includegraphics[width=0.48\textwidth]{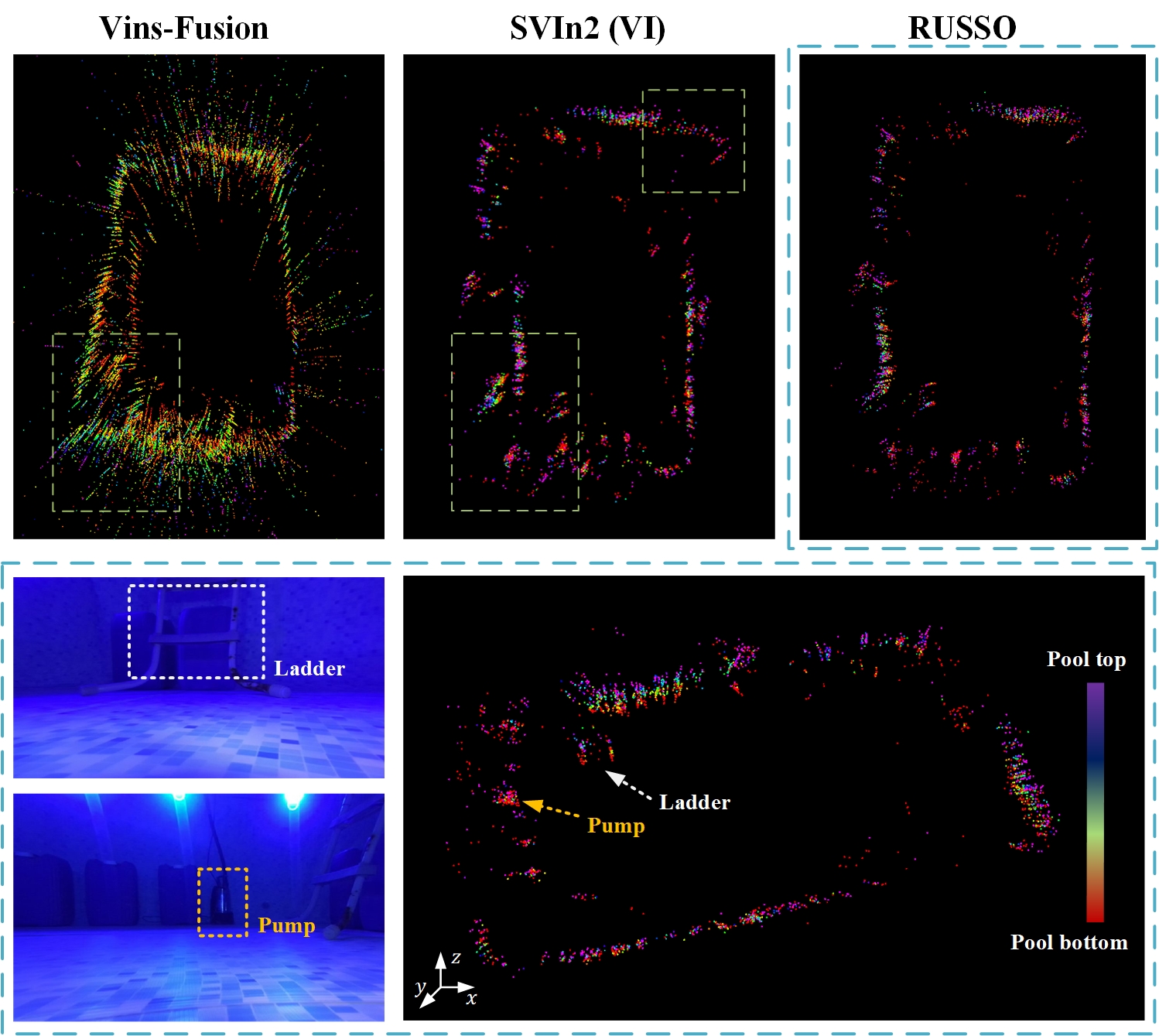}
  \caption{Comparison of mapping results for different SLAM algorithms in the rectangular lab pool. The first row presents the top-view results, while the second row shows an alternative view of the RUSSO map. The green dotted boxes highlight the areas of map drift. Arrows and dotted boxes of the same color indicate corresponding locations in the laboratory pool and the map.}
  \label{fig:Mapping}
\end{figure}

Additionally, RUSSO still shows better localization accuracy in cases when visual features are sufficient, as indicated by the localization errors in pool sequences 2 and 3 presented in Table \ref{tab:RMSE}. The maps constructed by the three different SLAM algorithms in pool sequence 3 are shown in Fig. \ref{fig:Mapping}. The map produced by RUSSO displays a rectangular structure that accurately represents the actual shape of our lab pool, whereas the maps from the other two methods appear inconsistent due to the accumulated drift.

\subsubsection{SLAM Initialization}
\begin{figure}[h]
  \centering
  \includegraphics[width=0.48\textwidth]{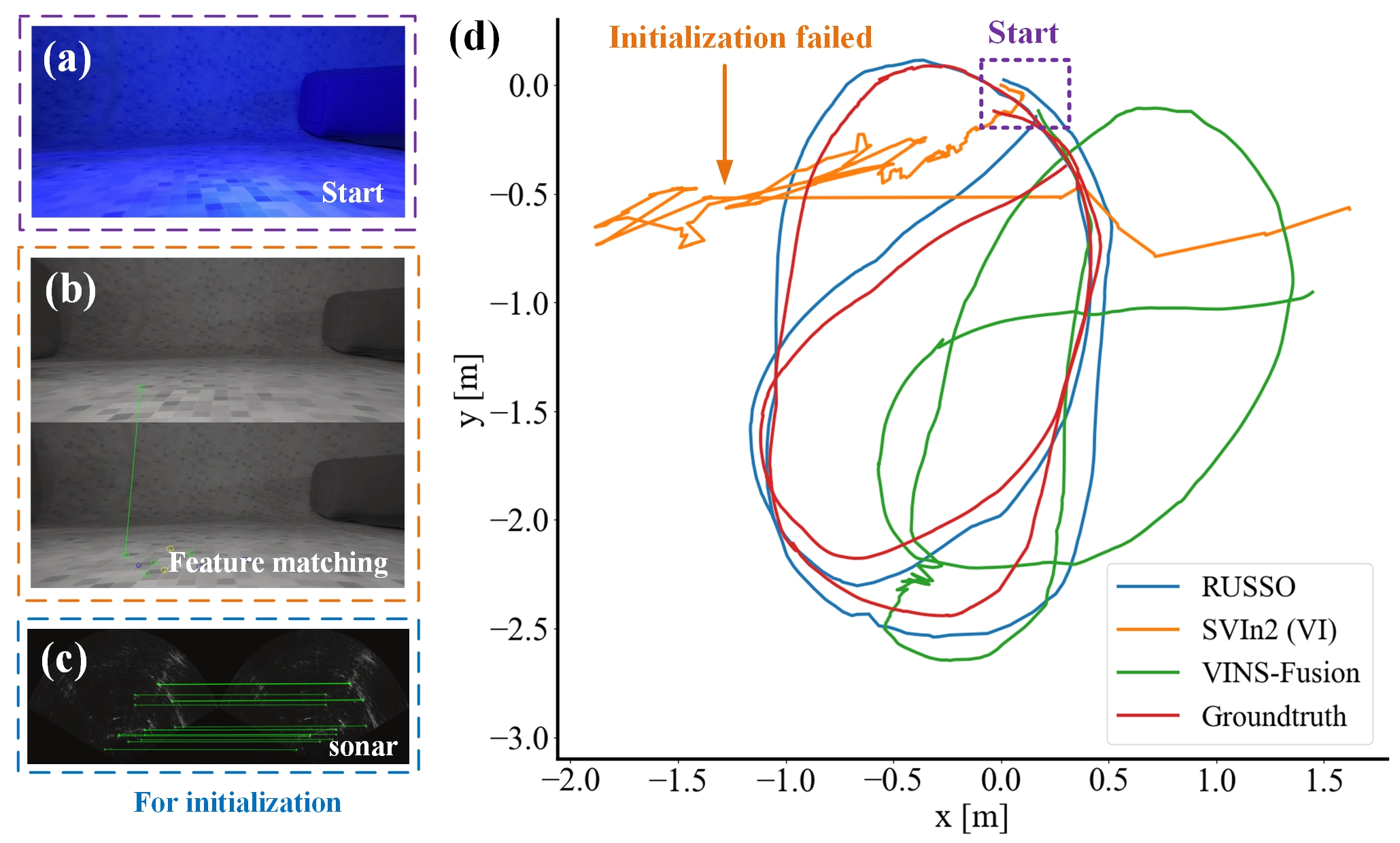}
  \caption{(a) The scene where SLAM experiments launched. (b) The result of feature matching in SVIn2. (c) The result of sonar matching in the launched scene. (d) The comparison of different SLAM algorithms with ground truth.}
  \label{fig:Initialization trajectory}
\end{figure}

To validate the effectiveness of our robust initialization strategy, we launch all SLAM algorithms in a visual degradation scene shown in Fig. \ref{fig:Initialization trajectory}(a). Due to the lack of visual features, as illustrated in Fig. \ref{fig:Initialization trajectory}(b), SVIn2 (VI) fails to initialize in Fig. \ref{fig:Initialization trajectory}(d). Although VINS-Fusion can initialize successfully, there is a large drift in the overall estimated trajectory due to inaccurate initialization. Quantitative results in Table \ref{tab:RMSE} show that RUSSO with good initialization leads to better localization accuracy than VINS-Fusion.

\begin{figure}[h]
  \centering
  \includegraphics[width=0.48\textwidth]{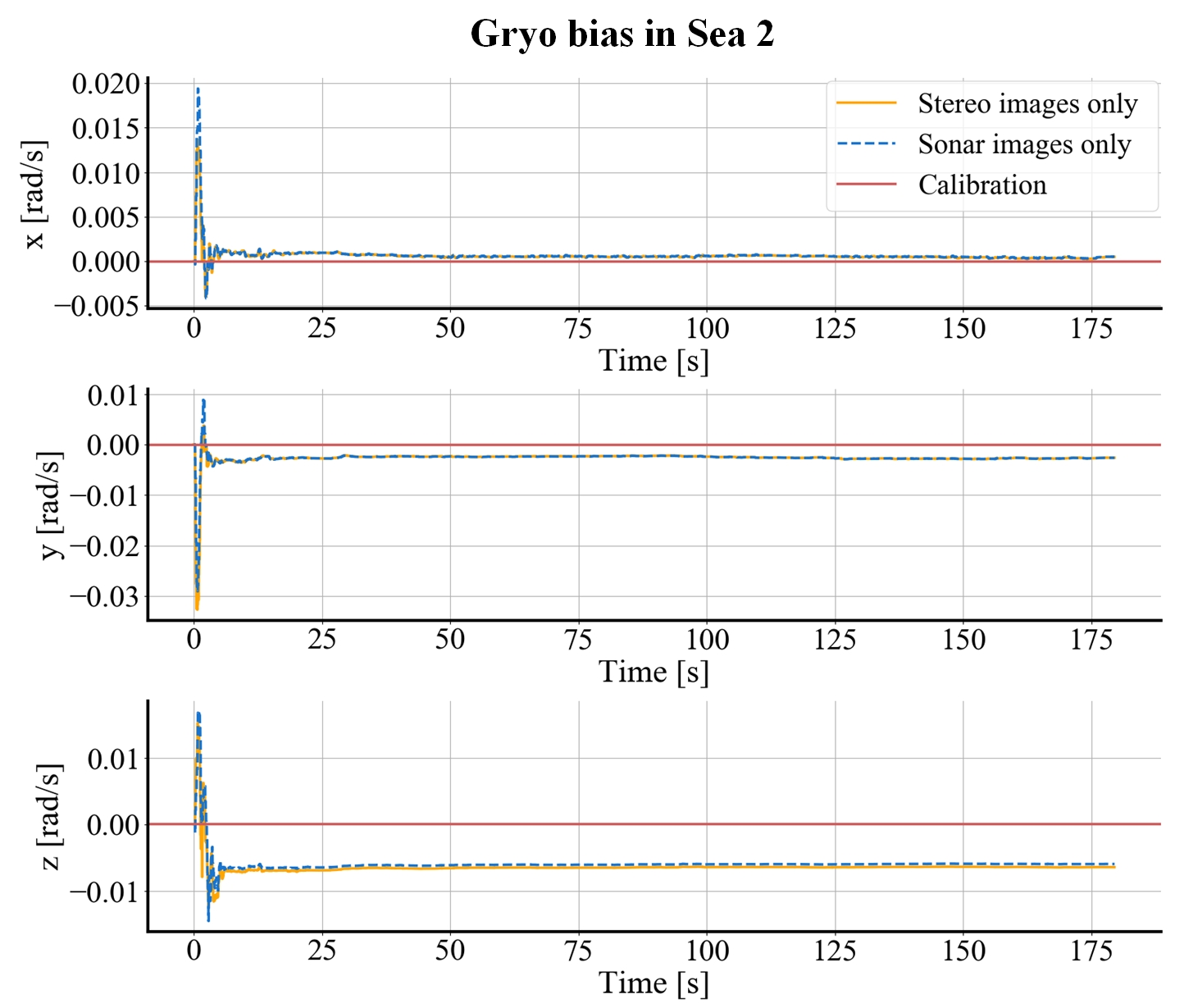}
  \caption{The gyroscope bias estimated by RUSSO using different types of data for initialization.}
  \label{fig:Initialization evaluation}
\end{figure}

We further conducted an experiment in sea sequence 2 that used two different types of data (stereo camera images or sonar images) for initialization respectively and evaluated the gyroscope bias estimation of the two approaches. The results are depicted in Fig. \ref{fig:Initialization evaluation}, which show that our proposed sonar initialization method can be as accurate as the one using stereo camera images for initialization. Nonetheless, the estimation with only sonar initialization is more jittery at the beginning due to the degradation of sonar elevation angles.

\begin{figure*}[t!]
    \centering
    \includegraphics[width=1\linewidth]{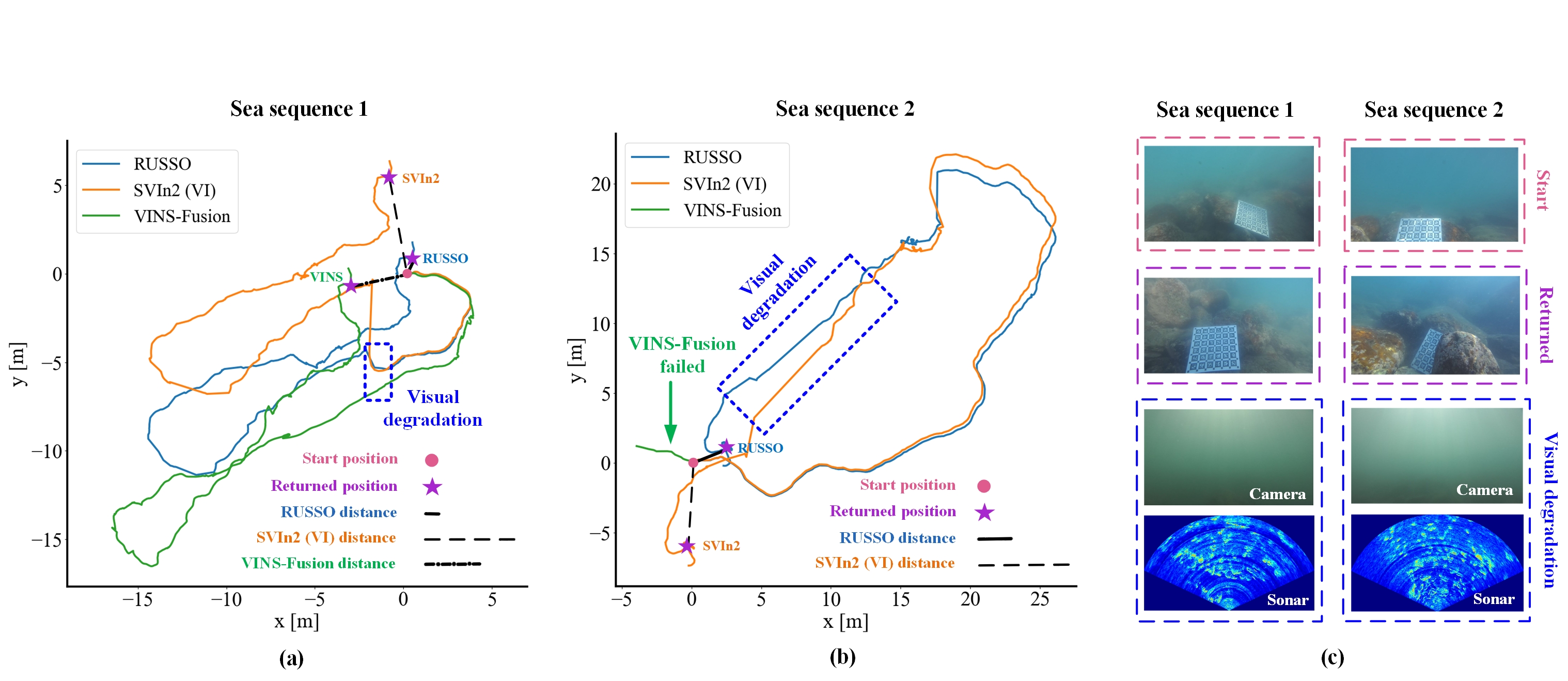}
    \caption{Results of the sea experiments. The red dotted box indicates the starting point of the robot, the purple dotted box represents the retured point of the robot, where the calibration board is visible within the camera's field of view. The blue dotted box shows the robot suffer from visual degradation. The black dotted and solid lines display the true distance between the start point and the returned point. (a) The trajectory results of Sea 1.  (b) The trajectory results of Sea 2. VINS-Fusion fails to initialize. (c) The images captured by the camera at the start and returned points in the two sea sequences, along with the images from the cameras and sonar during severe visual degradation.}
    \label{fig:sea trajectory}
\end{figure*}

\subsubsection{IMU Propagation Optimization } In RUSSO, we propose to use sonar pose estimation to provide a good prior for IMU propagation if it is under visual degradation.  Fig. \ref{fig:pool trajectory}(a) and (b) depict the estimated trajectories of RUSSO with and without the sonar prior. Table \ref{tab:pool_RMSE} quantitatively indicates that it improves localization accuracy by using sonar estimation to provide a better prior for IMU propagation when visual degradation occurs.

\begin{figure}[h]
  \centering
\includegraphics[width=0.48\textwidth]{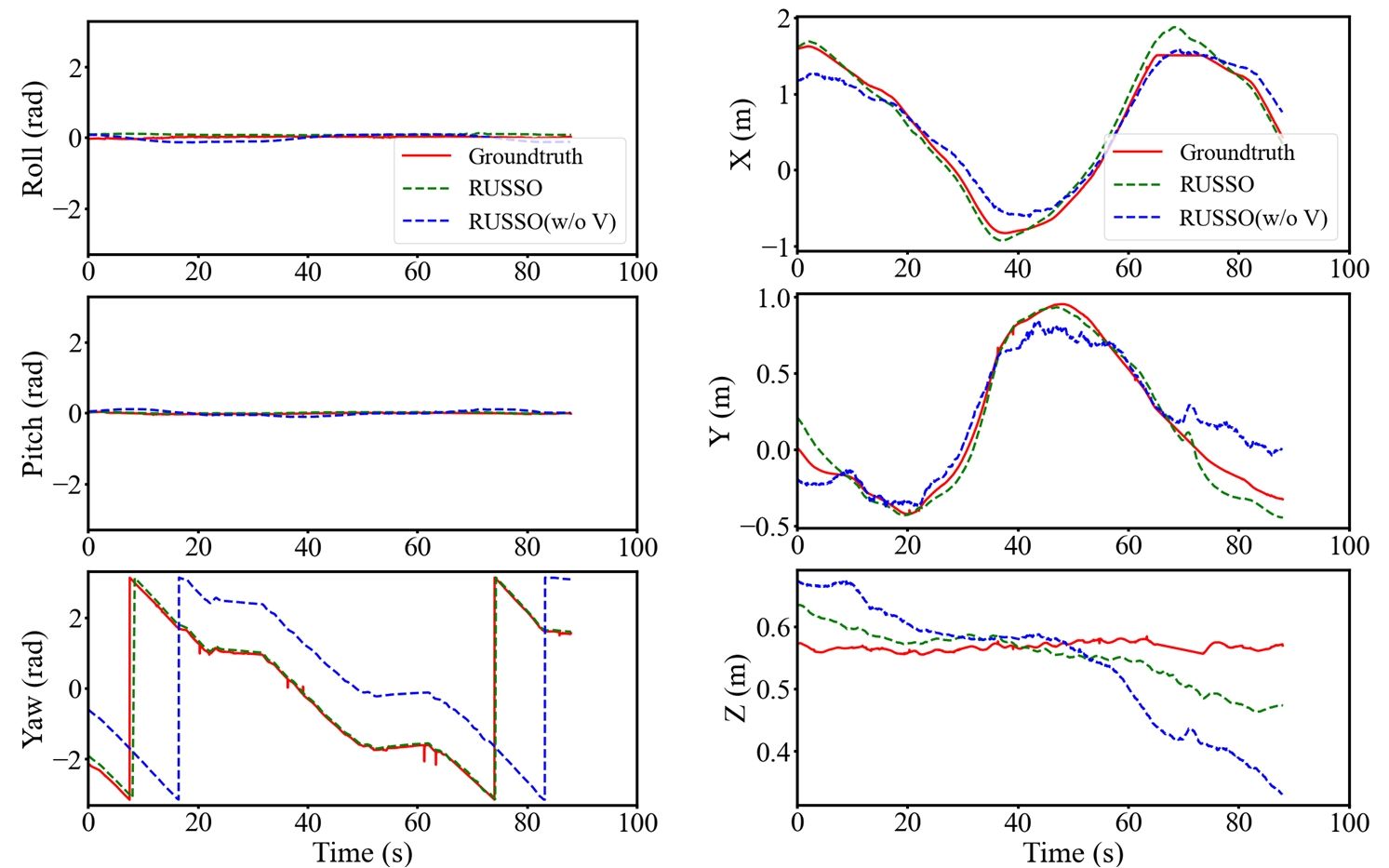}
  \caption{The comparison result for RUSSO with and without visual information. RUSSO (w/o V) means RUSSO without visual information.}
  \label{fig:XYZRPY}
\end{figure}

In the following experiment, we remove visual information from RUSSO. As shown in Fig. \ref{fig:XYZRPY}, RUSSO is reduced to a 3-DoF pose estimation system. Estimates on translation on the Z-axis and rotation on pitch and roll become unobservable, and RUSSO produces a larger error on these degree-of-freedom.

\subsection{Experiments in The Sea}
Furthermore, we perform experiments in the two  sea sequences which use the same robot platform as employed in the lab pool experiments, except using the high-precision Xsense IMU instead of the low-cost one in ZED2. In the sea experiment, we place a calibration board at the place where the robot starts and then control the robot to return to the starting point after making a loop around the sea, as depicted by the red dotted boxes and blue dotted boxes in Fig. \ref{fig:sea trajectory}(c).  But the loop closure module is always turned off for every experimented SLAM algorithm. The estimated trajectories of each SLAM algorithm in sea 1 and sea 2 are shown in Fig. \ref{fig:sea trajectory}(a), and (b). The red points are captured at the position where the sequence starts, and the blue points are from the returned position. 

In the sea experiments, it is difficult to construct a ground-truth by using COLMAP \cite{schonberger2016pixelwise} or other 3D mapping algorithms since these are too many visual degradation scenes during the entire sequence. To quantitatively analyze the localization error, we adopt the following steps:
\begin{table}[ht]
    \footnotesize
    \centering
    \caption{Quantitative result of pool 1}
    \label{tab:pool_RMSE}
    \scalebox{1}{
    \begin{tabular}{c|c|c}
    \hline
    Method    & Translation RMSE (m) & Rotation RMSE ($^{\circ}$) \\ \hline
    RUSSO & \textbf{0.208} & 2.063    \\ \hline
    RUSSO (w/o P) & 0.259 & 2.058   \\\hline
    SVIn2 (VI) & 0.353 & 2.054   \\\hline
    VINS-Fusion & 0.629 & 2.158\\\hline
    \end{tabular}
    }
\end{table}
\subsubsection{Calculate the relative pose using calibration board} 
\begin{figure}[h!]
  \centering
  \includegraphics[width=0.48\textwidth]{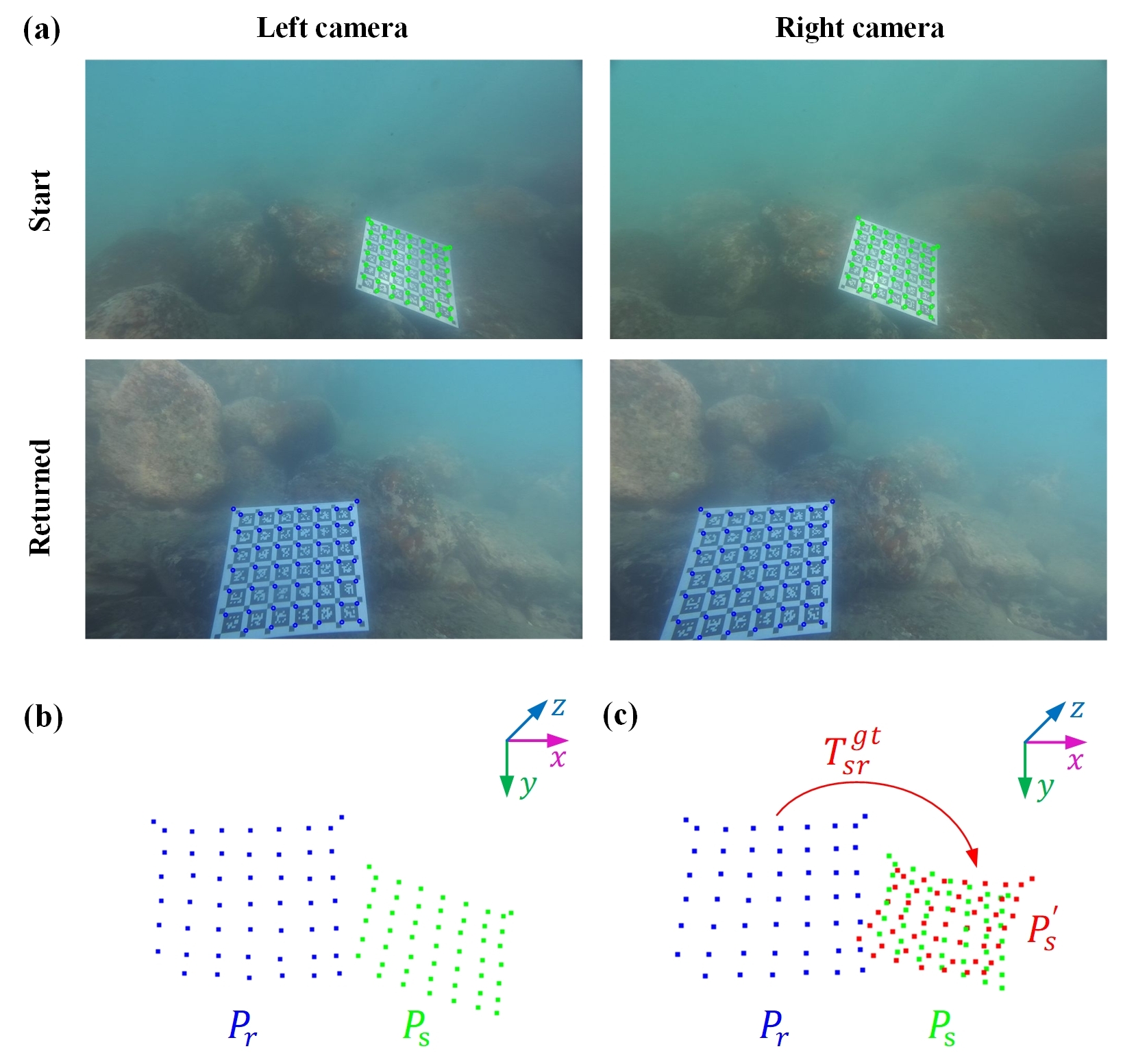}
  \caption{(a) Feature points are manually selected for binocular stereo on the stereo images for the start and returned frame. (b) The resulted 3D pointcloud via binocular stereo. The blue points are reconstructed using starting frame and the green points are reconstructed using the return frame. (c) The relative transformation  between the returned points and the start points.}
  \label{fig:Sea GT}
\end{figure}

Since the calibration board is visible both in the starting frame and in the end frame of the two sequences, we can calculate the relative pose transformation $\mathbf{T}^{gt}_{sr}$ between the returned and start position by looking at the control points on the calibration board, as shown in Fig. \ref{fig:Sea GT}. We validate the accuracy of the transformation by transforming returned points $\mathbf{P}_r$ to $\mathbf{P}^{'}_s$, as show in \ref{fig:Sea GT}(c). The mean distance between start points $\mathbf{P}_s$ and the transformed points $\mathbf{P}^{'}_s$ are 0.022 m, 0.013 m, 0.019 m in $X$, $Y$, $Z$ direction, respectively. Due to the presence of errors in camera calibration as well as errors in hand-selected feature points in pixels, we consider this transformation to be acceptable as the ground-truth.
\subsubsection{Calculate the localization error} We now have the ground-truth relative pose transformation $\mathbf{T}^{gt}_{sr}$, so we can calculate the relative pose error $\mathbf{e}^{RUSSO}_{rr}$, $ \mathbf{e}^{SVIn2}_{rr}$, $\mathbf{e}^{VINS}_{rr}$  by using the following formulation:
\begin{equation}
  \begin{bmatrix}
    \mathbf{e}^{RUSSO}_{rr} \\
    \mathbf{e}^{SVIn2}_{rr} \\
    \mathbf{e}^{VINS}_{rr} \\
  \end{bmatrix}
  =
  \begin{bmatrix}
    \mathbf{T}^{gt^{-1}}_{sr} * \mathbf{T}^{RUSSO}_{sr} \\
    \mathbf{T}^{gt^{-1}}_{sr} * \mathbf{T}^{SVIn2}_{sr} \\
    \mathbf{T}^{gt^{-1}}_{sr} * \mathbf{T}^{VINS}_{sr} \\
  \end{bmatrix}
\end{equation}
where $\mathbf{T}^{RUSSO}_{sr}$ is the relative pose between the starting frame and the returned frame of our estimation; $\mathbf{T}^{SVIn2}_{sr}$ is the relative pose of SVIn2 (VI) and $\mathbf{T}^{VINS}_{sr}$ the estimation from VINS-Fusion. The quantitative localization error in the two sea sequences are shown in Table \ref{tab:RMSE}.

In the sequence Sea 1, all three algorithms can run the full course,  SVIn2 (VI) encounters a large drift when visual degradation occurs, as shown in the blue dotted box in Fig. \ref{fig:sea trajectory}(a) and (c), with no features or texture in the camera image. The sonar image presents clear features at this moment, therefore RUSSO can leverage observation from sonar images for pose optimization and diminish the drift. VINS-Fusion yields a significant drift due to the visual degradation. The trajectory scale of VINS-Fusion is obviously larger than that of the other two algorithms, resulting in a large distance between the returned point and the starting point, as shown by the black dotted line in Fig. \ref{fig:sea trajectory}(a). It is clear that RUSSO achieves the best localization accuracy with 1.409m error after travelling more than 50m, while the error of SVIn2 is 3.667m and VINS-Fusion is 2.354m.

In the sequence Sea 2, VINS-Fusion fails to initialize since the  time the camera sees the calibration board is too short and VINS-Fusion requires fair amount of features for initialization. SVIn2 (VI) performs well in the first half of the sequence but it drifts when suffering visual degradation later, as shown by the blue dotted box in Fig. \ref{fig:sea trajectory}(b). In addition, the long period of visual degradation makes a large estimated distance between the returned and the starting position by SVIn2 (VI), as shown by the black dotted line. In contrast, RUSSO provides accurate pose estimation against visual degradation, as evidenced by the close distance between the estimated returned point and starting point.



\section{Conclusion}\label{sec: conclusion}
This work presents RUSSO, a robust underwater SLAM system that fuses imaging sonar with a stereo camera and IMU against visual degradation. A sonar odometry is first proposed, whose pose estimation further facilitates IMU propagation and SLAM initialization under visual degradation scenarios. We conduct extensive experiments in the simulator, pool, and shallow sea to verify the feasibility of RUSSO. The experiments demonstrate that RUSSO outperforms other two SOTA visual-inertial SLAM algorithms in all experimental scenarios. Our proposed method of using imaging sonar as a constraint in RUSSO effectively improves yaw
estimation accuracy in approximately fixed-depth underwater tasks such as underwater surveying and mapping.
How to extend this approach to full 6-DoF estimation remains an area for future research.


\bibliography{References}

\begin{thebibliography}{10}
\providecommand{\url}[1]{#1}
\csname url@rmstyle\endcsname
\providecommand{\newblock}{\relax}
\providecommand{\bibinfo}[2]{#2}
\providecommand\BIBentrySTDinterwordspacing{\spaceskip=0pt\relax}
\providecommand\BIBentryALTinterwordstretchfactor{4}
\providecommand\BIBentryALTinterwordspacing{\spaceskip=\fontdimen2\font plus
\BIBentryALTinterwordstretchfactor\fontdimen3\font minus \fontdimen4\font\relax}
\providecommand\BIBforeignlanguage[2]{{%
\expandafter\ifx\csname l@#1\endcsname\relax
\typeout{** WARNING: IEEEtran.bst: No hyphenation pattern has been}%
\typeout{** loaded for the language `#1'. Using the pattern for}%
\typeout{** the default language instead.}%
\else
\language=\csname l@#1\endcsname
\fi
#2}}

\bibitem{xanthidis2020navigation}
M.~Xanthidis, N.~Karapetyan, H.~Damron, S.~Rahman, J.~Johnson, A.~O’Connell, J.~M. O’Kane, and I.~Rekleitis, ``Navigation in the presence of obstacles for an agile autonomous underwater vehicle,'' in \emph{2020 IEEE International Conference on Robotics and Automation (ICRA)}.\hskip 1em plus 0.5em minus 0.4em\relax IEEE, 2020, pp. 892--899.

\bibitem{yang2022monocular}
P.~Yang, H.~Liu, M.~Roznere, and A.~Q. Li, ``Monocular camera and single-beam sonar-based underwater collision-free navigation with domain randomization,'' in \emph{The International Symposium of Robotics Research}.\hskip 1em plus 0.5em minus 0.4em\relax Springer, 2022, pp. 85--101.

\bibitem{9204397}
E.~Simetti, R.~Campos, D.~D. Vito, J.~Quintana, G.~Antonelli, R.~Garcia, and A.~Turetta, ``Sea mining exploration with an uvms: Experimental validation of the control and perception framework,'' \emph{IEEE/ASME Transactions on Mechatronics}, vol.~26, no.~3, pp. 1635--1645, 2021.

\bibitem{xu2021underwater}
S.~Xu, T.~Luczynski, J.~S. Willners, Z.~Hong, K.~Zhang, Y.~R. Petillot, and S.~Wang, ``Underwater visual acoustic slam with extrinsic calibration,'' in \emph{2021 IEEE/RSJ International Conference on Intelligent Robots and Systems (IROS)}.\hskip 1em plus 0.5em minus 0.4em\relax IEEE, 2021, pp. 7647--7652.

\bibitem{huang2023tightly}
Y.~Huang, P.~Li, S.~Yan, Y.~Ou, Z.~Wu, M.~Tan, and J.~Yu, ``Tightly-coupled visual-dvl fusion for accurate localization of underwater robots,'' in \emph{2023 IEEE/RSJ International Conference on Intelligent Robots and Systems (IROS)}.\hskip 1em plus 0.5em minus 0.4em\relax IEEE, 2023, pp. 8090--8095.

\bibitem{rahman2018sonar}
S.~Rahman, A.~Q. Li, and I.~Rekleitis, ``Sonar visual inertial slam of underwater structures,'' in \emph{2018 IEEE International Conference on Robotics and Automation (ICRA)}.\hskip 1em plus 0.5em minus 0.4em\relax IEEE, 2018, pp. 5190--5196.

\bibitem{johannsson2010imaging}
H.~Johannsson, M.~Kaess, B.~Englot, F.~Hover, and J.~Leonard, ``Imaging sonar-aided navigation for autonomous underwater harbor surveillance,'' in \emph{2010 IEEE/RSJ International Conference on Intelligent Robots and Systems}.\hskip 1em plus 0.5em minus 0.4em\relax IEEE, 2010, pp. 4396--4403.

\bibitem{9286733}
M.~Teng, L.~Ye, Z.~Yuxin, J.~Yanqing, Z.~Qianyi, and A.~M. Pascoal, ``Efficient bathymetric slam with invalid loop closure identification,'' \emph{IEEE/ASME Transactions on Mechatronics}, vol.~26, no.~5, pp. 2570--2580, 2021.

\bibitem{mcconnell2022overhead}
J.~McConnell, F.~Chen, and B.~Englot, ``Overhead image factors for underwater sonar-based slam,'' \emph{IEEE Robotics and Automation Letters}, vol.~7, no.~2, pp. 4901--4908, 2022.

\bibitem{mcconnell2020fusing}
J.~McConnell, J.~D. Martin, and B.~Englot, ``Fusing concurrent orthogonal wide-aperture sonar images for dense underwater 3d reconstruction,'' in \emph{2020 IEEE/RSJ International Conference on Intelligent Robots and Systems (IROS)}.\hskip 1em plus 0.5em minus 0.4em\relax IEEE, 2020, pp. 1653--1660.

\bibitem{burnett2024continuous}
K.~Burnett, A.~P. Schoellig, and T.~D. Barfoot, ``Continuous-time radar-inertial and lidar-inertial odometry using a gaussian process motion prior,'' \emph{arXiv preprint arXiv:2402.06174}, 2024.

\bibitem{mur2015orb}
R.~Mur-Artal, J.~M.~M. Montiel, and J.~D. Tardos, ``Orb-slam: a versatile and accurate monocular slam system,'' \emph{IEEE transactions on robotics}, vol.~31, no.~5, pp. 1147--1163, 2015.

\bibitem{wang2017stereo}
R.~Wang, M.~Schworer, and D.~Cremers, ``Stereo dso: Large-scale direct sparse visual odometry with stereo cameras,'' in \emph{Proceedings of the IEEE International Conference on Computer Vision}, 2017, pp. 3903--3911.

\bibitem{campos2021orb}
C.~Campos, R.~Elvira, J.~J.~G. Rodr{\'\i}guez, J.~M. Montiel, and J.~D. Tard{\'o}s, ``Orb-slam3: An accurate open-source library for visual, visual--inertial, and multimap slam,'' \emph{IEEE Transactions on Robotics}, vol.~37, no.~6, pp. 1874--1890, 2021.

\bibitem{leutenegger2015keyframe}
S.~Leutenegger, S.~Lynen, M.~Bosse, R.~Siegwart, and P.~Furgale, ``Keyframe-based visual--inertial odometry using nonlinear optimization,'' \emph{The International Journal of Robotics Research}, vol.~34, no.~3, pp. 314--334, 2015.

\bibitem{sun2018robust}
K.~Sun, K.~Mohta, B.~Pfrommer, M.~Watterson, S.~Liu, Y.~Mulgaonkar, C.~J. Taylor, and V.~Kumar, ``Robust stereo visual inertial odometry for fast autonomous flight,'' \emph{IEEE Robotics and Automation Letters}, vol.~3, no.~2, pp. 965--972, 2018.

\bibitem{kim2013real}
A.~Kim and R.~M. Eustice, ``Real-time visual slam for autonomous underwater hull inspection using visual saliency,'' \emph{IEEE Transactions on Robotics}, vol.~29, no.~3, pp. 719--733, 2013.

\bibitem{rahman2019SVIn2}
S.~Rahman, A.~Q. Li, and I.~Rekleitis, ``Svin2: An underwater slam system using sonar, visual, inertial, and depth sensor,'' in \emph{2019 IEEE/RSJ International Conference on Intelligent Robots and Systems (IROS)}.\hskip 1em plus 0.5em minus 0.4em\relax IEEE, 2019, pp. 1861--1868.

\bibitem{vargas2021robust}
E.~Vargas, R.~Scona, J.~S. Willners, T.~Luczynski, Y.~Cao, S.~Wang, and Y.~R. Petillot, ``Robust underwater visual slam fusing acoustic sensing,'' in \emph{2021 IEEE International Conference on Robotics and Automation (ICRA)}.\hskip 1em plus 0.5em minus 0.4em\relax IEEE, 2021, pp. 2140--2146.

\bibitem{hu2022tightly}
C.~Hu, S.~Zhu, Y.~Liang, and W.~Song, ``Tightly-coupled visual-inertial-pressure fusion using forward and backward imu preintegration,'' \emph{IEEE Robotics and Automation Letters}, vol.~7, no.~3, pp. 6790--6797, 2022.

\bibitem{raaj20163d}
Y.~Raaj, A.~John, and T.~Jin, ``3d object localization using forward looking sonar (fls) and optical camera via particle filter based calibration and fusion,'' in \emph{OCEANS 2016 MTS/IEEE Monterey}.\hskip 1em plus 0.5em minus 0.4em\relax IEEE, 2016, pp. 1--10.

\bibitem{yang2020extrinsic}
D.~Yang, B.~He, M.~Zhu, and J.~Liu, ``An extrinsic calibration method with closed-form solution for underwater opti-acoustic imaging system,'' \emph{IEEE Transactions on Instrumentation and Measurement}, vol.~69, no.~9, pp. 6828--6842, 2020.

\bibitem{lindzey2021extrinsic}
L.~Lindzey and A.~Marburg, ``Extrinsic calibration between an optical camera and an imaging sonar,'' in \emph{OCEANS 2021: San Diego--Porto}.\hskip 1em plus 0.5em minus 0.4em\relax IEEE, 2021, pp. 1--8.

\bibitem{norman2023actag}
K.~Norman, D.~Butterfield, and J.~G. Mangelson, ``Actag: Opti-acoustic fiducial markers for underwater localization and mapping,'' in \emph{2023 IEEE/RSJ International Conference on Intelligent Robots and Systems (IROS)}.\hskip 1em plus 0.5em minus 0.4em\relax IEEE, 2023, pp. 9955--9962.

\bibitem{alcantarilla2013fast}
P.~Alcantarilla, J.~Nuevo, and A.~Bartoli, ``Fast explicit diffusion for accelerated features in nonlinear scale spaces british machine vision conference (bmvc),'' 2013.

\bibitem{shin2015bundle}
Y.-S. Shin, Y.~Lee, H.-T. Choi, and A.~Kim, ``Bundle adjustment from sonar images and slam application for seafloor mapping,'' in \emph{OCEANS 2015-MTS/IEEE Washington}.\hskip 1em plus 0.5em minus 0.4em\relax IEEE, 2015, pp. 1--6.

\bibitem{lupton2011visual}
T.~Lupton and S.~Sukkarieh, ``Visual-inertial-aided navigation for high-dynamic motion in built environments without initial conditions,'' \emph{IEEE Transactions on Robotics}, vol.~28, no.~1, pp. 61--76, 2011.

\bibitem{qin2017vins}
T.~Qin, P.~Li, and S.~Shen, ``Vins-mono: A robust and versatile monocular visual-inertial state estimator,'' \emph{IEEE Transactions on Robotics}, vol.~34, no.~4, pp. 1004--1020, 2018.

\bibitem{zhang2022dave}
M.~M. Zhang, W.-S. Choi, J.~Herman, D.~Davis, C.~Vogt, M.~McCarrin, Y.~Vijay, D.~Dutia, W.~Lew, S.~Peters, \emph{et~al.}, ``Dave aquatic virtual environment: Toward a general underwater robotics simulator,'' in \emph{2022 IEEE/OES Autonomous Underwater Vehicles Symposium (AUV)}.\hskip 1em plus 0.5em minus 0.4em\relax IEEE, 2022, pp. 1--8.

\bibitem{manhaes2016uuv}
M.~M.~M. Manh{\~a}es, S.~A. Scherer, M.~Voss, L.~R. Douat, and T.~Rauschenbach, ``Uuv simulator: A gazebo-based package for underwater intervention and multi-robot simulation,'' in \emph{OCEANS 2016 MTS/IEEE Monterey}.\hskip 1em plus 0.5em minus 0.4em\relax IEEE, 2016, pp. 1--8.

\bibitem{pisano1998contrast}
E.~D. Pisano, S.~Zong, B.~M. Hemminger, M.~DeLuca, R.~E. Johnston, K.~Muller, M.~P. Braeuning, and S.~M. Pizer, ``Contrast limited adaptive histogram equalization image processing to improve the detection of simulated spiculations in dense mammograms,'' \emph{Journal of Digital imaging}, vol.~11, pp. 193--200, 1998.

\bibitem{schonberger2016pixelwise}
J.~L. Sch{\"o}nberger, E.~Zheng, J.-M. Frahm, and M.~Pollefeys, ``Pixelwise view selection for unstructured multi-view stereo,'' in \emph{Computer Vision--ECCV 2016: 14th European Conference, Amsterdam, The Netherlands, October 11-14, 2016, Proceedings, Part III 14}.\hskip 1em plus 0.5em minus 0.4em\relax Springer, 2016, pp. 501--518.

\end{thebibliography}
\bibliographystyle{IEEEtran}
\begin{IEEEbiography}
[{\includegraphics[width=1in,height=1.25in,clip,keepaspectratio]{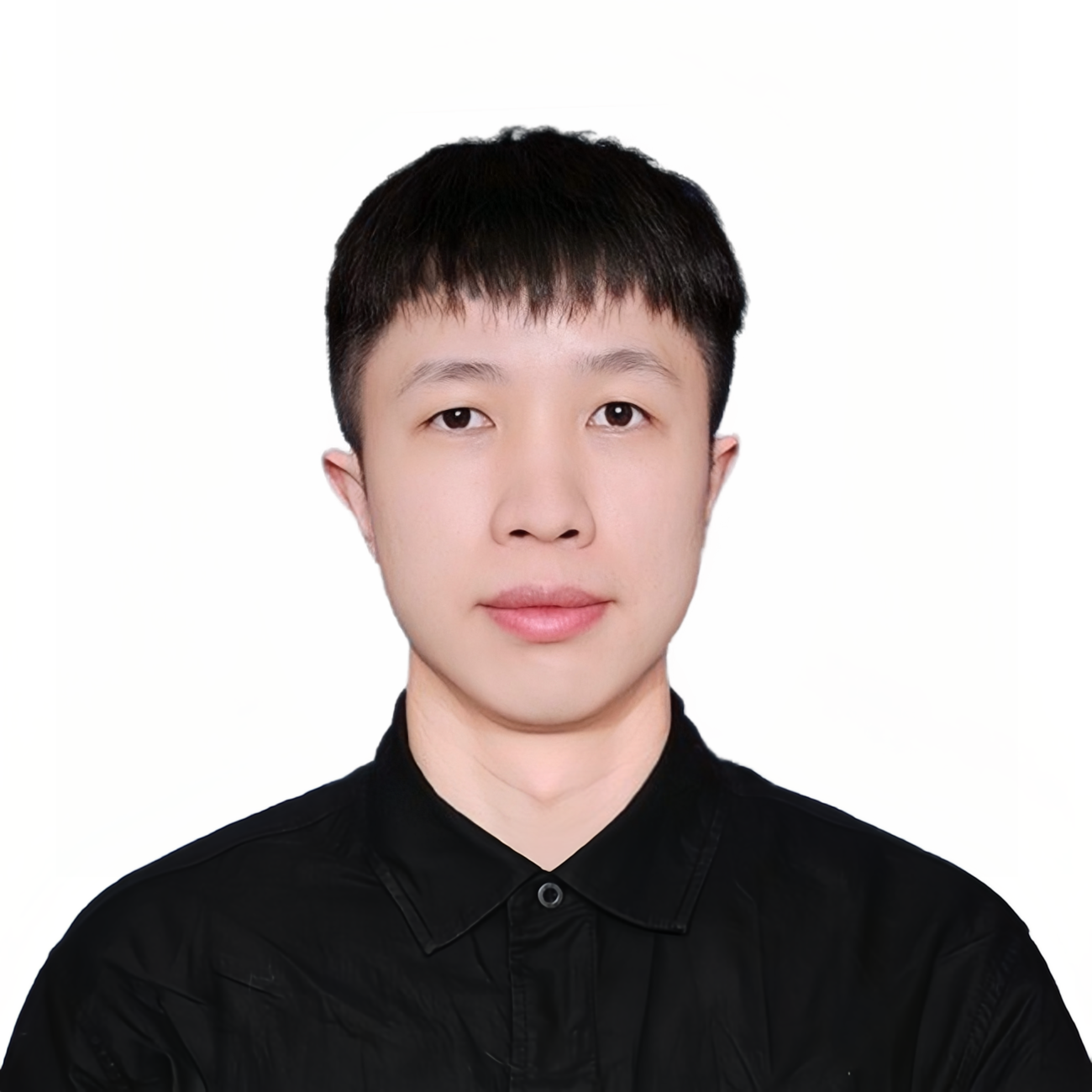}}]{Shu Pan}
received his B.S. degree in Electronic Information Science and Technology from Hainan Tropical Marine Institute, Hainan, China in 2018, and his M.S. degree in Electronic Information from South China Normal University, Guangdong, China in 2023. He is a research associate with Harbin Institute of Technology, Shenzhen, China. His research interests include localization and mapping of underwater robotics.
\end{IEEEbiography}
\vspace{-10 mm} 
\begin{IEEEbiography}
[{\includegraphics[width=1in,height=1.25in,clip,keepaspectratio]{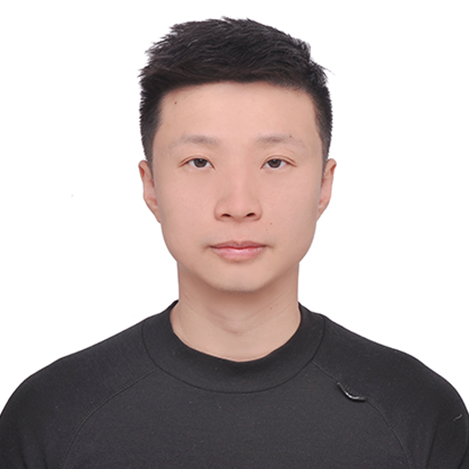}}]{Ziyang Hong}
received his B.S. degree in Electronic Information and Engineering from Shantou University, China in 2016, and his M.S and Ph.D. degrees in Vision and Robotics from Heriot-Watt University, UK in 2018 and 2023 respectively. He is currently a research associate with Harbin Institute of Technology, Shenzhen, China. His research interests include robotic vision and SLAM for mobile robotics.
\end{IEEEbiography}
\vspace{-10 mm} 
\begin{IEEEbiography}
[{\includegraphics[width=1in,height=1in,clip,keepaspectratio]{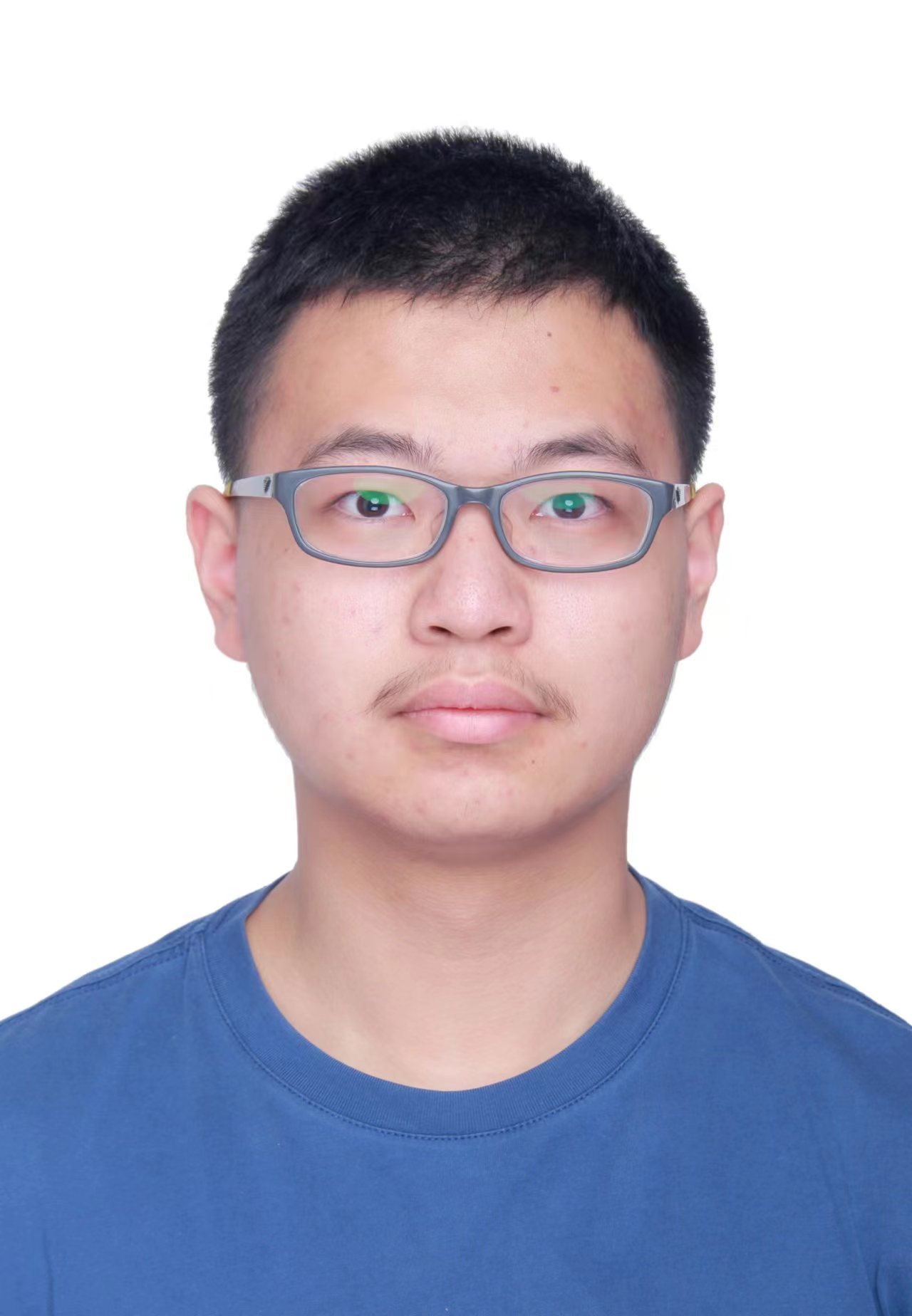}}]{Zhangrui Hu} currently is an undergraduate student majoring in Automation, in the School of Mechanical Engineering and Automation at Harbin Institute of Technology, Shenzhen, China. His research interests include localization and mapping of underwater robotics.
\end{IEEEbiography}
\vspace{-10 mm} 
\begin{IEEEbiography}
[{\includegraphics[width=1in,height=1.25in,clip,keepaspectratio]{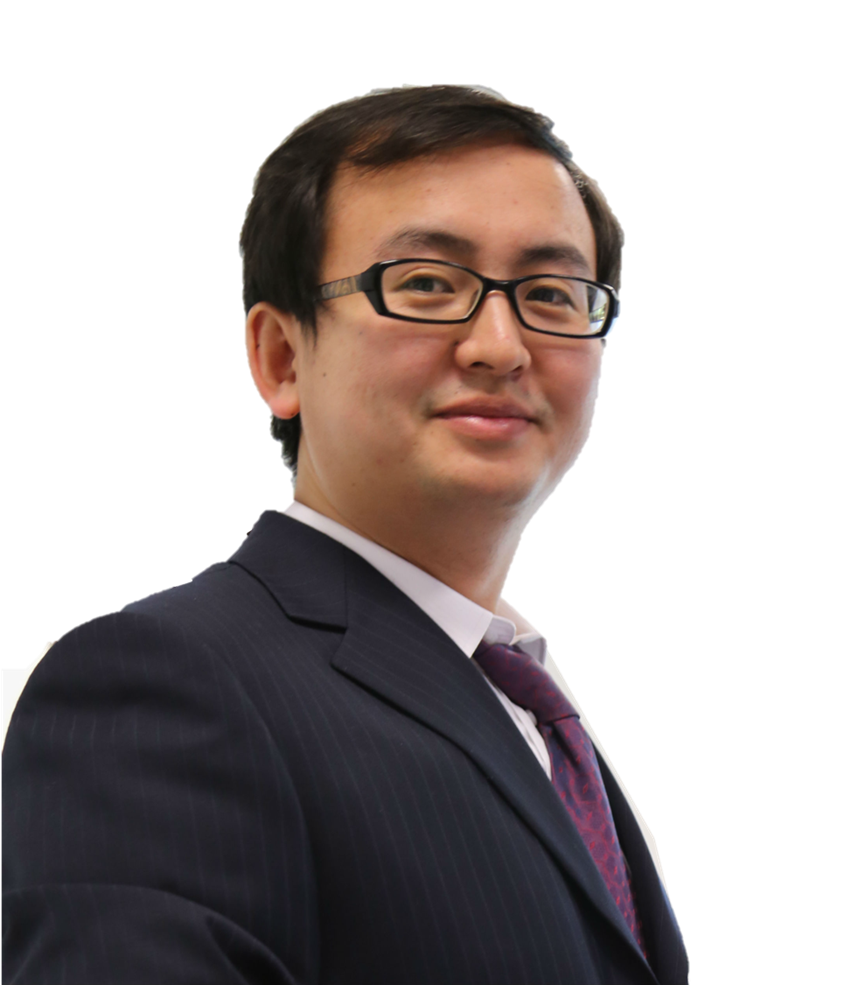}}]{Xiandong Xu} is an Associate Professor at Tianjin University. He received his B.S. and Ph.D. degrees in electrical engineering from Tianjin University, Tianjin, China, in 2009 and 2015. From 2015 to 2020, he was with Queen's University Belfast and Cardiff University as a post-doctoral researcher. His research focuses on integrated energy systems, in particular industrial demand response and offshore energy utilization. 
\end{IEEEbiography}
\vspace{-10 mm} 
\begin{IEEEbiography}
[{\includegraphics[width=1in,height=1.25in,clip,keepaspectratio]{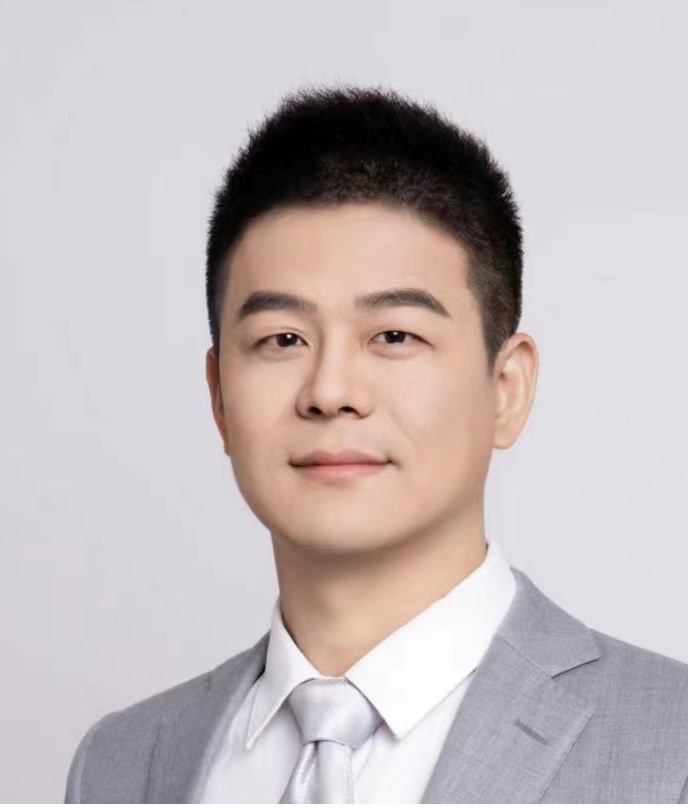}}]{Wenjie Lu}
received his B.S. in Mechatronic Engineering from Zhejiang University, Hangzhou, China in 2009. He received his M.S. and Ph.D. in Mechanical Engineering from Duke University, Durham, NC, USA in 2011 and 2014 respectively. He is currently an assistant professor with HIT (Shenzhen). His research interests include learning efficient motion planners and controllers for autonomous systems subject to uncertainties in 3D complex environments. 
\end{IEEEbiography}
\vspace{-10 mm} 
\begin{IEEEbiography}
[{\includegraphics[width=1in,height=1.25in,clip,keepaspectratio]{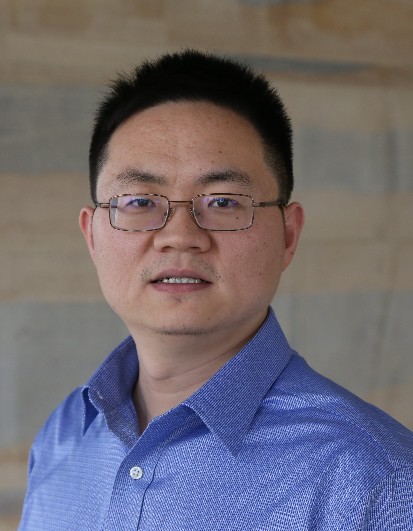}}]{Liang Hu}
received both the B.E. and M.E. degrees in Control Science and Engineering from Harbin Institute of Technology, China, in 2008 and 2010, respectively, and the Ph.D. degree in Computer Science from Brunel University London, UK, in 2016. 
He is a professor at the School of Mechanical Engineering and Automation at Harbin Institute of Technology, Shenzhen, China.  His research interests include localisation, navigation, control and their applications in robotics and autonomous systems.
\end{IEEEbiography}
\end{document}